\documentclass[lettersize,journal]{IEEEtran}
\usepackage{amsmath,amsfonts}
\usepackage{algorithm}
\usepackage{array}
\usepackage[caption=false,font=normalsize,labelfont=sf,textfont=sf]{subfig}
\usepackage{textcomp}
\usepackage{stfloats}
\usepackage{url}
\usepackage{verbatim}
\usepackage{graphicx}
\usepackage{color}

\usepackage{mathtools}
\usepackage[only,llbracket,rrbracket]{stmaryrd}

\usepackage{makecell}
\usepackage{xtab,afterpage}
\usepackage{tabularray}
\usepackage{amsmath}
\usepackage{verbatim} 
\usepackage{lipsum}
\usepackage{multirow}
\usepackage{rotating}
\usepackage{placeins}
\usepackage{tikz}
\usepackage{pgfplots}
\usetikzlibrary{calc,positioning}

\usepackage{algpseudocode}
\usepackage[numbers]{natbib}

\newcommand{\Cai}[1]{\textcolor{red}{\textbf{/* #1 (Cai) */}}}

\begin{document}

\title{Towards Causal Classification: A Comprehensive Study on Graph Neural Networks}  

\author{Simi~Job, 
        Xiaohui~Tao,~\IEEEmembership{Senior~Member,~IEEE},        
        Taotao Cai,
        Lin Li,
        Haoran Xie,~\IEEEmembership{Senior~Member,~IEEE},
        Jianming Yong
        


\thanks{*Corresponding authors: Simi Job (email: \textit{simi.job@unisq.edu.au}) and Xiaohui Tao (email: \textit{Xiaohui.Tao@unisq.edu.au}) are with School of Mathematics, Physics, and Computing, University of Southern Queensland, Australia}
\thanks{Taotao Cai is with School of Mathematics, Physics, and Computing, University of Southern Queensland, Australia (e-mail: \textit{Taotao.Cai@unisq.edu.au}).}
\thanks{Lin Li is with School of Computer Science and Artificial Intelligence, Wuhan University of Technology, China, (e-mail: \textit{cathylilin@whut.edu.cn).}}
\thanks{Haoran Xie is with Department of Computing and Decision Sciences, Lingnan University, Hong Kong (e-mail: \textit{hrxie@ln.edu.hk}).}
\thanks{Yong is with the School of Business, University of Southern Queensland, Springfield (e-mail: jianming.yong@unisq.edu.au).}
}


\markboth{Journal of \LaTeX\ Class Files,~Vol.~14, No.~8, August~2021}%
{Shell \MakeLowercase{\textit{et al.}}: A Sample Article Using IEEEtran.cls for IEEE Journals}


\maketitle

\begin{abstract}
The exploration of Graph Neural Networks (GNNs) for processing graph-structured data has expanded, particularly their potential for causal analysis due to their universal approximation capabilities. Anticipated to significantly enhance common graph-based tasks such as classification and prediction, the development of a causally enhanced GNN framework is yet to be thoroughly investigated. Addressing this shortfall, our study delves into nine benchmark graph classification models, testing their strength and versatility across seven datasets spanning three varied domains to discern the impact of causality on the predictive prowess of GNNs. This research offers a detailed assessment of these models, shedding light on their efficiency, flexibility in different data environments, and highlighting areas needing advancement. Our findings are instrumental in furthering the understanding and practical application of GNNs in diverse data-centric fields.
\end{abstract}

\begin{IEEEkeywords}
Graph, Causality, Causal Learning, Graph Neural Networks, Classification
\end{IEEEkeywords}


\section{Introduction}\label{sec:introduction}

Causal knowledge is the understanding of cause and effect relationships among various factors, extending beyond mere correlation. It focuses on comprehending the interactions between elements that result in specific outcomes and explores how changes in one aspect can impact another element within a system. Recently, there has been an increased focus on exploring causality, with researchers acknowledging the importance of incorporating causal knowledge into data modeling. Causality has found numerous applications in several domains such as economics \cite{xie2022forest}, social sciences \cite{ridley2020poverty}, medicine \cite{brouwers2020non} and healthcare \cite{prosperi2020causal}, environmental science \cite{adebayo2022environmental} etc. For instance, in domains like medicine, causality can explore factors that affect treatment. Likewise, one can investigate the causal factors that elevate the risk of medical conditions. In social sciences, the causal factors that contribute to economic inequalities may be studied. Hence, integrating causal knowledge is highly advantageous and pertinent in contemporary studies.

In the context of graph-structured data, causality refers to modeling the relationships of cause and effect among variables represented in a graph. Graph Neural Networks (GNN) are neural networks that can process and infer graph-structured data. GNNs have been used in various domains including recommendation \cite{fan2019graph, chang2021sequential}, urban intelligence \cite{li2022graph}, medicine \cite{gharsallaoui2021investigating} and healthcare \cite{sharma2023ai}, community detection \cite{luo2021detecting}, fraud detection \cite{li2023lgm} and so on. GNNs have demonstrated remarkable performance in various tasks such as node or graph classification tasks and link prediction tasks. Nevertheless, GNNs have few limitations including over-smoothing, interpretability and generalizability problems, sensitivity to graph structure and limited ability in capturing long-range dependencies. In recent times, GNNs are being explored for causal analysis due to their ability to function as universal approximators of data \cite{zevcevic2021relating}. Introducing causality to GNN architecture can substantially address the aforementioned limitations of GNNs. By examining the inherent causal relationships within the data, it becomes possible to prioritize relevant information. This approach helps mitigate the issue of over-smoothing to a certain extent. The causal links between distant nodes can capture long-range information and the cause-effect relationships also promote extraction of transferable features leading to generalizability. Moreover, causal relationships are less sensitive to structure changes. 

\begin{figure}[h]
    \centering
    \includegraphics[width=\columnwidth]{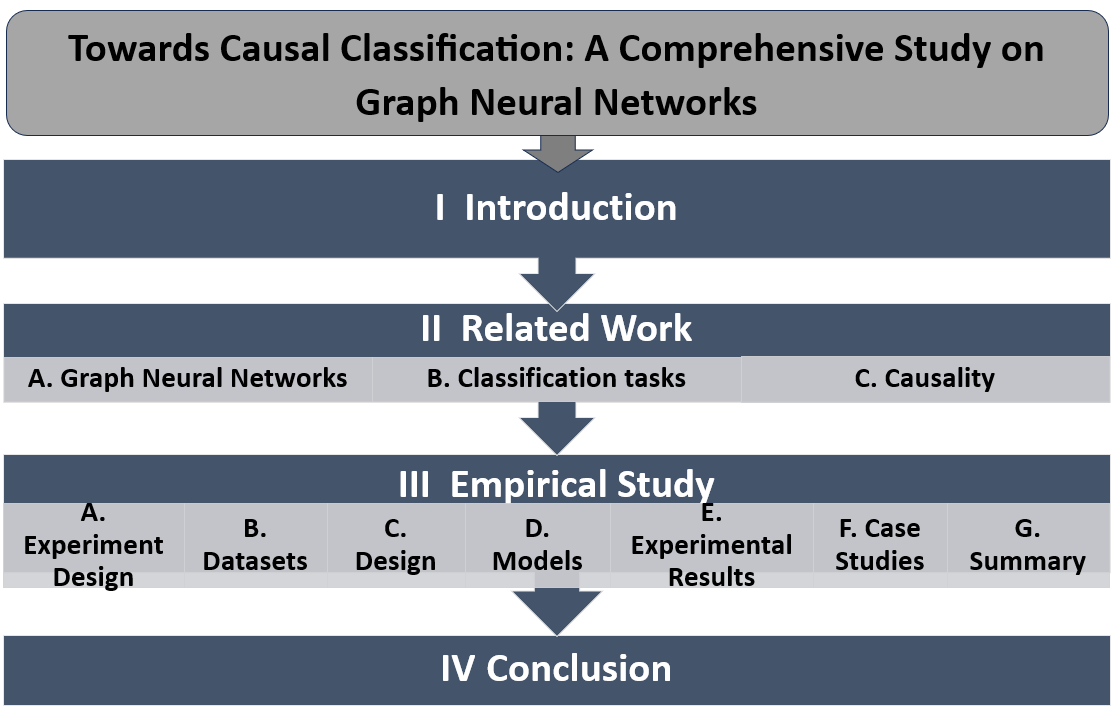}
    \caption{An Overview of the paper}
    \label{fig:overview}
\end{figure}

Through this study, we aim to conduct a thorough investigation into the application of graph neural networks for classification tasks. Few studies similar to our work are the benchmark studies by \cite{dwivedi2020benchmarking}, which focused on graph positional encoding in GNNs and by \cite{li2022emerging}, which studied GNNs for fault diagnosis. \cite{kosan2023gnnx} conducted a benchmarking study that focused on GNN explainers. An in-depth study by \cite{chen2022bag} researched deep GNN architectures and experimented on various citation network datasets with various model settings. None of these studies have researched causality for GNN classification and hence this work is proposed to fill the research gap in analysing the significance of causality in graph prediction models. This study aims to conduct a comprehensive study on the most representative models that are used in graph neural networks classification tasks, with the potential of incorporating causal elements into the respective frameworks. Fig.~\ref{fig:overview} depicts an overview of our paper illustrating the key components of this work.

\subsection*{Contributions}
Our research offers a thorough examination and insights into the capabilities and applications of GNNs. The key contributions of this study include:
\begin{itemize}
    \item \textbf{Comprehensive Analysis:} A detailed study on the application and performance of graph neural networks in both graph and node classification tasks.
    \item \textbf{Causal Relationship Evaluation:} A critical investigation into the causal capabilities of five representative models, contrasting their performance with four baseline GNN models.
    \item \textbf{Investigation of potential causal approaches:} Researching models based on attention mechanisms and mutual information estimation for their potential in causal classification. 
    \item  \textbf{Cross-domain evaluation:}  Evaluation of the models using datasets from three different domains for exploring their generalizability.
    \item \textbf{Multi-class graph classification:} Additional instances of graph classification cases for further evaluation using larger and multi-class datasets.
\end{itemize}

The remainder of the paper is organized as follows: Section~\ref{sec-LR} provides an overview of research studies related to our work. Section~\ref{sec-ES} presents the empirical study and discusses the experimental results. Section~\ref{sec-CON} concludes the paper with a brief summary of our work.

\section{Related Work}\label{sec-LR}

This section reviews existing literature focusing on graph neural networks (GNNs) and their variants, causality, and their applications in classification tasks.

\subsection{Graph Neural Networks}

Graph Neural Networks (GNNs) are a type of neural networks that can process graph structured data and are capable of capturing data dependencies through message passing between nodes. A node is represented in terms of its features and the features of its neighbouring nodes. The most basic GNN is the Graph Convolutional Networks (GCN) \cite{KipfThomasN2017SCwG}. GCN employ convolution operations to aggregate features including neighbourhood features. During convolution, information is propagated through nodes and subsequently pooling layers are used for graph pooling. The convolutional layer captures the local graph structure by means of this aggregation operation. Following convolution, an activation function is applied and the final output layer is used to produce the specific task-related predictions. GraphSAGE (Graph SAmple and aggreGatE) \cite{hamilton2017inductive} introduces inductive learning to the GCN framework, where sampling and aggregation of features from a node's neighborhood are used to build node representations. GraphSAGE samples a subset of local neighbors for each node and subsequently aggregates this information. This approach facilitates large-scale graph processing. Graph Attention Networks (GAT) \cite{VeličkovićPetar2018GAN} incorporates attention mechanism to the GNNs and can capture complex dependencies in graph data. Attention scores are computed for node pairs, from which attention coefficients are obtained. Weighted sum is used to aggregate neighbourhood information, which enables the model to capture the most relevant information. Graph Isomorphism Networks (GIN)
GINs \cite{xu2018powerful} are GNNs that are specifically designed for the task of determining whether two graphs are structurally identical. GINs employ a customised aggregate function that render them invariant to node ordering in a graph. 

GNNs learns node embeddings for capturing structural information in a graph. \cite{you2019position} proposed Position-aware Graph Neural Networks (P-GNNs) for capturing the position of a node in a graph. Position-aware node embeddings are computed by sampling sets of anchor nodes and estimating the distance between the target and anchor nodes. The node positional information is expected to boost the performance of GNNs in diverse tasks such as link prediction and node classification. A data augmented GNN model called the GAUG graph data augmentation
framework was proposed by \cite{zhao2021data} for the purpose of improving semi-supervised node classification. Neural edge predictors are used as a means of exposing GNNs to likely edges and limiting exposure to unlikely ones. A GCN framework called Stacked and Reconstructed Graph Convolutional Networks for Recommender Systems (STAR-GCN) by \cite{zhang2019star} stacked GCN encoder-decoders for learning node representations. They used intermediate supervision and reconstructed masked input node embeddings to generate embeddings for new nodes. A temporal GCN called T-GCN was proposed by \cite{zhao2019t} for traffic prediction which used GCN for learning spatial dependencies and GRU for capturing temporal dependencies of traffic networks, with nodes representing roads and edges representing road connections. Similarly, a GAT-based spatio temporal framework called ST-GAT \cite{song2022st} employed attention mechanism for traffic speed prediction. The model incorporated individual spatial and temporal dependencies using Individual Spatio-Temporal graph (IST-graph) and Spatio-Temporal point (ST-point) embedding. A self-attention mechanism is employed to learn patterns hidden in
IST-dependencies among these ST-points for accurate embeddings. Graph Isomorphism Network (GIN) was used for drug–drug interaction (DDI) predictions with DDIGIN proposed by \cite{wang2023improved}, with the model using Node2Vec for obtaining initial representations. These representations are then optimized by aggregation of first-order neighboring information from graphs. The GIN framework is purported to improve the expressive power of representations.

\subsection{Classification tasks}

Classification tasks in the graph domain remains a formidable challenge in the field of machine learning and the emergence of GNNs has significantly advanced the capabilities of learning graph representations and managing large-scale graph datasets. Classification tasks include node-level or graph-level tasks. Graph-level classification involves predicting class labels of the graph as a whole. On the other hand, node-level classification involves predicting class labels of individual nodes in a graph. For an input graph $G = (V,E) $, where $V$ and $E$ are sets of nodes and edges respectively, the output is a class label $y$ indicating the class of $G$. In the context of graph classification, GNNs compute node embeddings by iteratively updating them through the aggregation of neighborhood embeddings. The final embeddings are subjected to the pooling layer for aggregating as a single embedding which is passed to the final classification layer.

A degree-specific GNN framework called DEMO-Net was designed by \cite{wu2019net} for node and graph classification, towards incorporating structure-aware neighbourhoods and a degree-aware framework for classification tasks. The model was influenced by the Weisfeiler-Lehman graph isomorphism test to identify 1-hop neighbourhood structures. \cite{maurya2022simplifying} investigated the node feature aggregation step in node classification to design Feature Selection Graph Neural Network (FSGNN) for extracting relevant features. The study discovered that there are less informative features that can affect prediction performance and equipped FSGNN to learn the most relevant features. \cite{wang2022minority} proposed minority-weighted GNN (mGNN) for extracting information from imbalanced data, particularly in the context of social network analysis. The approach addresses the problem of imbalanced classification, with a focus on node classification. The problem of unattributed node classification was researched by \cite{sun2022generalized}, who proposed a generalized equivariance property and a Preferential Labeling technique for addressing this issue. The former permits additional auto-isomorphic permutation and the latter technique achieves the generalized equivariance property asymptotically. The framework proves particularly advantageous in addressing real-world challenges such as anonymized social networks. A GIN-based model called  Dynamic Multi-Task Graph Isomorphism Network (DMT-GIN) proposed by \cite{wang2023dynamic} transformed fMRI images into brain network structures for classification of Alzheimer’s disease. DMT-GIN incorporated attention mechanism for capturing node features and graph structural information.

\subsection{Causality}

Causality entails understanding the cause and/or effect relationships between different elements in data. Hence, causal relationships form a critical factor in understanding data, rendering the concept of causality pertinent to any machine learning task. Causality is an efficient tool for feature selection and has been investigated recently for its potential to extract relevant features. \cite{yu2020causality} developed a causality-based feature selection package called CausalFS encompassing the most representative algorithms in this domain. The methods included constraint-based algorithms and score-based approaches with Markov Boundary (MB) learning in different scenarios such as simultaneous MB learning, Divide-and-conquer MB learning and MB learning  with relaxed assumptions. Causal representation learning \cite{scholkopf2021toward} involves discovering causal variables from raw observations and is a challenging task in causal learning. The function approximation capabilities of GNNs serve to model nonlinear causal relations in large-scale graph data. Moreover, causality plays a crucial role in improving feature selection and hence is critical for GNN classification tasks. Causal models also provide insights into the underlying relationships in graph structures and hence contribute to model interpretability. 

A causal attention learning model called CAL was proposed by \cite{sui2022causal} for graph classification. The model discovers causal patterns in data through attention mechanism and employs an attention-based GNN for this purpose. A similar approach was used by \cite{wang2023causal} to build Causal-Trivial Attention Graph Neural Network (CTA-GNN) for discovering causality patterns by diminishing confounding effects of shortcut features. CTA-GNN was employed in fault diagnosis of complex industrial processes, wherein the industrial system entities were modelled as nodes, with their interactions represented as edges.

Mutual information (MI) may also be useful for deriving causal relationships from graphs. For instance, high MI can indicate a potential causal relationship, subject to other influential factors in the data. Unsupervised Hierarchical Graph Representation (UHGR) was proposed by \cite{ding2020unsupervised} for classification tasks using MI. The model was based on MI maximization between global and local parts for learning structural information. \cite{di2020mutual} also used MI maximization for classification tasks and link prediction tasks using GNNs. The authors accomplished MI maximization by neighbourhood enlargement in GNN aggregation. They further verified the model's reliability through experiments on datasets from diverse domains.

\section{Empirical Study} \label{sec-ES}

An empirical study was conducted with nine representative models employed in graph classification tasks and the details are discussed in this section. These models were chosen through a thorough examination of current literature, considering their potential for future exploration in the realm of causality-oriented Graph Neural Networks (GNNs). 
The outcomes of the empirical investigation are anticipated to probe the applicability of the algorithms across diverse settings and domains, aiming to illustrate both the strengths and constraints of the chosen models. Building upon these discoveries, researchers can further explore and devise innovative algorithms for causal classification utilizing Graph Neural Networks.

\subsection{Experiment Design}

Our study endeavors to investigate prominent Graph Neural Network algorithms used in classification, aiming to identify architectures with substantial potential for integrating causality. The research questions are carefully formulated to specifically examine the performance of Graph Neural Networks in classification tasks, with a goal of improving generalizability and accuracy rates, and understanding the role of hyperparameter tuning. The particular focus lies on assessing the potential improvements in classification performance when causality is incorporated. Addressing these research questions would provide future researchers with opportunities to identify mechanisms for extracting causal relationships from data, subsequently utilizing them to build more resilient models for enhanced classification performance. The research questions (\textit{RQ}) and the corresponding hypotheses (\textit{H}) explored in this study are as follows: 

\begin{itemize}
    \item \textbf{RQ1.} Which GNN architectures consistently exhibit the highest classification accuracy across all datasets? \\
    \textit{H.} The causality-enabled \textit{GAT-CAL} architecture with its attention mechanism is expected to demonstrate superior classification accuracy compared to other architectures, specifically the non-GAT models.

    \item \textbf{RQ2.} How do the baseline GNN architectures compare with the enhanced GNN models in terms of performance and generalizability across domains? \\
    \textit{H.} Although baseline GNN models such as GCN may exhibit better computational performance, enhanced GNN models are expected to demonstrate superior classification performance and generalizability. This is attributed to their capability to integrate relevant features.
    
    \item \textbf{RQ3.} Are there models that are particularly exceptional in specific tasks, such as node classification, in terms of performance?\\
    \textit{H.} GraphSAGE is expected to be the most suitable model for node classification, since it samples and aggregates information from the neighborhood of each node for learning representations.

     \item \textbf{RQ4.} Are the GNN architectures sensitive to hyperparameters and can the model performance improve with hyperparameter tuning? \\
    \textit{H.} Hyperparameter tuning is expected to hold the potential to improve classification performance, especially for certain GNN architectures.


\end{itemize}

  \begin{table}[hbt!]
\caption{Summary of datasets used in the study }
 \centering
  \begin{tabular}
  {|p{1.3cm}|p{0.6cm}|p{0.6cm}|p{1.2cm}|p{0.6cm}|p{1.2cm}|p{0.4cm}|}
 
  \hline
     & & & & & & \\
    
    \textbf{Dataset}  &  \textbf{\# }  & \textbf{ Avg.}  & \textbf{ \# nodes} & \textbf{Avg.} & \textbf{\# Edges} & \textbf{\# } \\
     & \textbf{graphs} & \textbf{nodes} & (1\textsuperscript{st} graph) & \textbf{edges} & (1\textsuperscript{st} graph) & \textbf{class} \\
  \hline
Cora  & 1 & - & 2708 & - & 10556 & 7 \\
  Citeseer   & 1 & - & 3327 & - & 9104 & 6 \\
  NCI1  & 4110 & 29.87 & 21 & 32.30 & 42 & 2 \\
  Proteins & 1113 & 39.06 &  42 & 72.82 & 162 &   2 \\
  Mutag   & 188 & 17.93 & 17 & 19.79 & 38 & 2 \\
   IMDB-B   & 1000 & 	19.77 & 20 & 96.53 & 146 & 2 \\
   REDDIT-B  & 2000 & 429.63 & 218 & 497.75 & 480 & 2 \\
   
          \hline
       
      \end{tabular}
\label{tab:tabldatasets}
\end{table}

 \subsection{Datasets} \label{sec-dataS}

This section describes the datasets that are used for experimental studies in this research. The datasets are selected from three diverse domains viz. Biochemical, citation and social networks, with the purpose of generalizing the research framework. These datasets are publicly available and are easily accessible. Moreover, these datasets have been widely used in similar studies, making them well-suited for reproducibility and comparability with similar models. The summary of the datasets are given in Tab.~\ref{tab:tabldatasets}.

\subsubsection*{Bio-Chemical datasets}

\begin{itemize}
    \item NCI1 \cite{wale2008comparison} is a cheminformatics dataset where each graph represents a chemical compound. The nodes represent atoms in a molecule and the edges represent bonds between atoms.
    
    \item Proteins \cite{borgwardt2005protein} is a proteins dataset with nodes representing amino acids and belong to two classes, enzymes or non- enzymes. Two nodes are connected with edges if they are less than 6 angstrom in distance.
    
    \item Mutag \cite{debnath1991structure} dataset consists of nitroaromatic compounds, with nodes representing atoms and the edges representing bonds between atoms. The dataset has two classes according to their mutagenic effect on a bacterium. The main limitation of this dataset is its modest size, although this aspect can be advantageous for conducting preliminary experiments with novel algorithms.

    These datasets, which are made available through the TUDataset package, are commonly used as benchmarks in the graph classification domain, facilitating the comparison of different graph-based algorithms. Moreover, they consist of molecular structures derived from real-world compounds, enhancing its applicability to practical scenarios. Proteins and NCI1 have been used by \cite{ding2020unsupervised}, with all three datasets employed for classification tasks in studies including \cite{xu2018powerful, sui2022causal, zhou2023deep, xie2022active}.
\end{itemize}

\subsubsection*{Citation Network datasets}

\begin{itemize}
    \item Cora \cite{McCallumAndrewKachites2000AtCo} is a citation network dataset consisting of scientific publications categorised into seven classes, with nodes and edges representing paper and citation relationships respectively. 
    \item Citeseer \cite{10.1145/276675.276685} is a citation network dataset consisting of scientific publications categorised into six classes, with nodes and edges representing paper and citation relationships respectively.

These two datasets are commonly used benchmark datasets and are representative of real citation networks and have been used for classification tasks in several studies \cite{velivckovic2017graph, kipf2016semi, ding2020unsupervised}. They possess a moderate size, enabling researchers to conduct extensive graph-based experiments. These datasets have few limitations, including the absence of temporal information and the need for improved feature representations. 

\end{itemize}

\subsubsection*{Social Network datasets} 

\begin{itemize}
    \item IMDB-BINARY (IMDB-B) \cite{yanardag2015deep} is a movie collaboration dataset consisting of movie information from IMDB. The nodes in a graph represent actors/actresses, with an edge between nodes for actors from same movie. The dataset contains collaboration graphs based on genres, with ego-networks for each individual. The graph is labelled based on movie genres viz. Romance or Action. This dataset has been used for classification tasks by \cite{sui2022causal}. However, due to its limitation to the movie genre, there is a need to investigate the dataset's generalizability to other domains.

    \item  REDDIT-BINARY (REDDIT-B) \cite{yanardag2015deep} consists of data related to online discussion threads, where nodes represent users and an edge between two nodes denote that a correspondence (comments) has been made between these two users. The graph is labelled based on whether it belongs to a question/answer-based or a discussion-based community.

    Few advantages of these datasets are their real-world relevance and the availability of a substantial amount of labeled data for training. Furthermore, these two datasets are widely utilized in graph-based research, facilitating comparative studies \cite{xu2018powerful, zhou2023multi,moon2023subgraph}.
\end{itemize}

 \subsection{Design} \label{sec-EDesign}

\begin{figure}[hbt!]
    \centering
    \includegraphics[width=2.5in,height=3in]{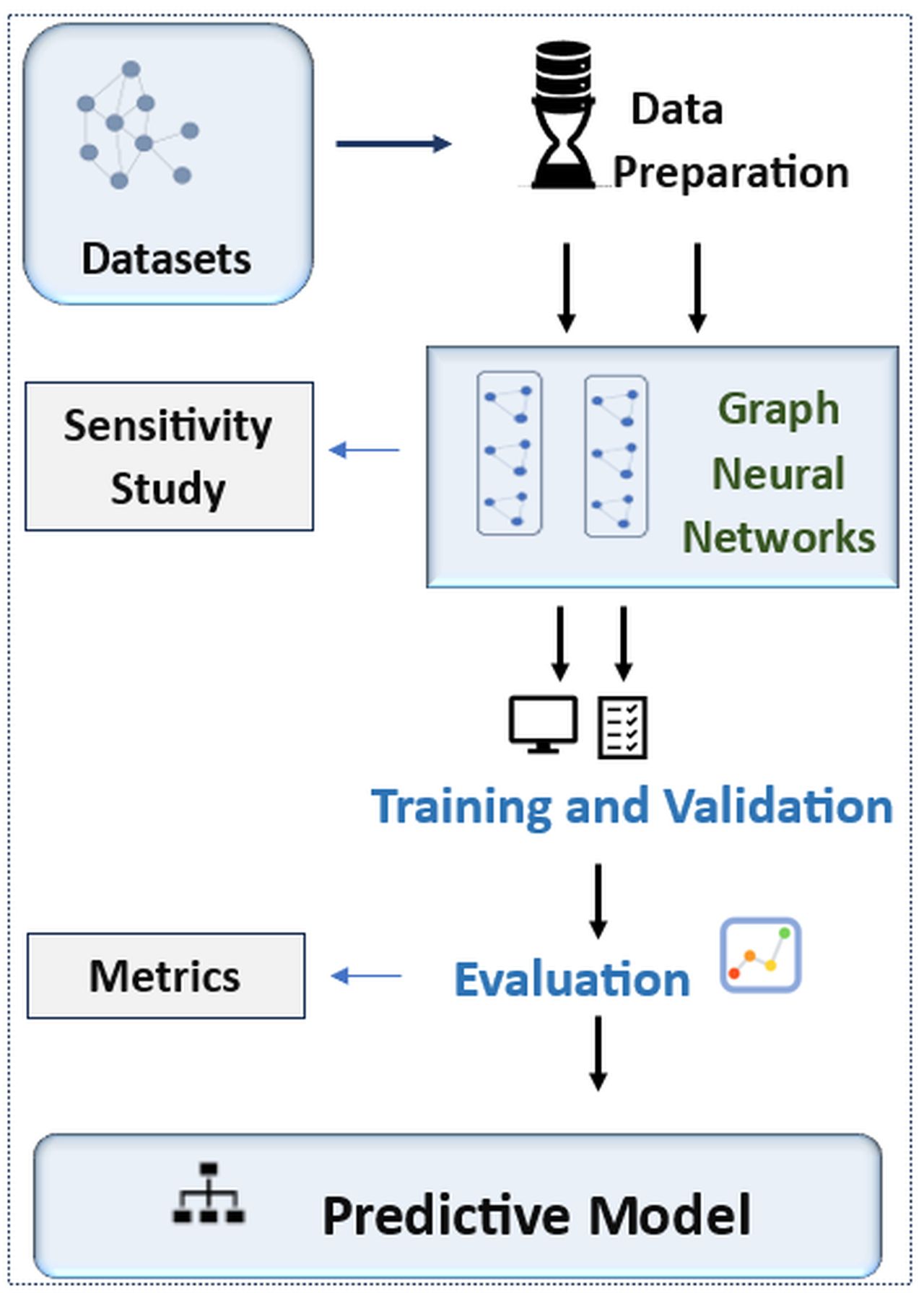}
    \caption{Task flow for the Experiment Design}
    \label{fig:taskflow}
\end{figure}

The experiments are designed based on the workflow illustrated in Fig.~\ref{fig:taskflow}. The graph datasets are split into training and testing sets, wherein KFold is used to generate indices for splitting the data into folds. A 5-fold cross validation is performed for assessing the performance and generalization ability of the models. A GNN model is designed for each respective architecture such as GCN, GAT etc., featuring a final classification layer. The model is then trained and evaluated using the standard performance metrics such as accuracy, precision/recall and F-scores. These metrics were selected based on their capacity to evaluate different aspects of a model's classification performance. While accuracy delivers a thorough evaluation of a model's correctness across all classes, its efficacy diminishes in imbalanced datasets. In such cases, precision, recall, and F-scores become crucial for a more nuanced assessment. Precision quantifies the accuracy of positive predictions, whereas recall measures the capacity to capture all positive instances. The F1-score balances precision and recall into a single metric. Additionally, a sensitivity study is carried out to investigate how variations in hyperparameters affect the performance of the model. 

 \subsection{Models} \label{sec-Algos}
 
The most representative and the most extensively employed models in GNN-based studies were investigated and subsequently, nine relevant models were selected for this study as follows:

\begin{itemize}
    \item GCN \cite{kipf2016semi}: The research primarily focuses on graph neural networks for classification tasks and hence the most commonly used GNN architecture viz. GCN is used as a baseline model. Comparing the fundamental structure of GCN with more advanced versions of GNNs would aid in determining whether enhanced GNN variants offer a substantial improvement in performance. GCN utilizes the convolutional mechanism to propagate information across a graph by aggregating features from neighboring nodes. GCN is highly effective in capturing local graph structures.
    
    \item GAT \cite{velivckovic2017graph}: GAT is a state-of-the-art GNN model that employs an attention mechanism, allowing nodes to selectively prioritize different neighbors when aggregating information. In contrast to GCN, GAT excels in capturing long-range dependencies, making it well-suited for tasks that necessitate the capture of both local and global patterns within the graph. With its attention mechanism, GAT assigns adaptive weights to the neighbors of each node during information aggregation. GAT can dynamically adjust neighbourhood importance based on learned attention weights and hence is advantageous in graph classification tasks.
    
    \item GIN \cite{xu2018powerful}: GIN is permutation invariant and hence can address the over-smoothing problems encountered by GCN. GIN is most typical for processing isomorphic graphs and is researched for its flexibility in capturing complex graph patterns. GIN consists of isomorphism layers that are invariant to node ordering. The node features are aggregated using summation allowing the model to capture relationships within a graph. GIN, characterized by its isomorphic and permutation invariant properties, has considerable potential in effectively capturing intricate structural patterns present within graphs. Therefore, it is crucial to explore the model's capabilities in the context of graph classification rather than focusing primarily on node classification.
    
    \item GraphSAGE \cite{hamilton2017inductive}: GraphSAGE involves node sampling followed by aggregation of the sampled node features using an aggregation function such as mean aggregation or pooling aggregation. With its localized sampling and aggregation approach, the model can efficiently handle large graphs. Moreover, GraphSAGE exhibits insensitivity to node ordering, making it a key contender as a model for tasks related to graph classification.
    GraphSAGE is investigated for its ability to capture more localized and diverse information from the graph. GraphSAGE is also the state-of-the-art model in efficiently adapting to diverse graph types.
    
    \item GCN-CAL, GAT-CAL, GIN-CAL \cite{sui2022causal}: The CAL framework was selected for its use of causality in GNN classification. The framework introduced causal attention learning to GCN, GAT and GIN architectures for enabling causal classification. The model primarily used a GNN-based encoder for obtaining node representations, followed by utilization of two MLPs for estimating edge and node-level attention scores. Leveraging attention scores is a fundamental approach for extracting causality through the use of Graph Neural Network (GNN) architectures. Therefore, this model is being investigated  for its performance in tasks related to causal classification.
    
    \item UHGR-GAT, UHGR-GCN  \cite{ding2020unsupervised}: The UHGR framework uses an encoder for constructing node representation, followed by graph pooling. The UHGR models use mutual information and hence we investigate this method for their ability to derive causality from graph data. A discriminator module is used for training the encoder for mutual information maximization. The learned graph representations are employed for node and graph classification tasks. In this work, two UHGR variants, using GAT and GCN, are studied for their potential for mutual information-based classification. 

\end{itemize}

The experiments on the selected models were conducted on the seven datasets described in Section~\ref{sec-dataS}. The selected models are further explained here. \cite{sui2022causal} proposed Causal Attention Learning (CAL) with mitigation of confounding effects using softmask estimation from attention scores. The graph is decomposed to causal and trivial attended graphs with two GNN layers. The authors proposed disentanglement of causal and trivial features, with GNNs filtering shortcut patterns for capturing causal features. The CAL framework was experimented with GCN, GAT and GIN architectures and for these three models, the settings used by \cite{sui2022causal} were mostly reproduced for experimentation. A 5-fold cross-validation is conducted, deviating from the 10-fold cross-validation used in the initial study. Though the original study involved training the models for 100 epochs, this study trains for 20 epochs due to minimal impact on the results. Similar settings were also used for the GCN, GAT, GIN and GraphSAGE models.

A graph representation learning framework called Unsupervised Hierarchical Graph Representation (UHGR) was proposed by \cite{ding2020unsupervised} for classification tasks. The model was based on mutual information (MI) and used MI maximization between global and local parts to learn related structural information. Encoder modules using GCN and GAT were used for UHGR-GCN and UHGR-GAT respectively to encode node representations. \cite{ding2020unsupervised} trained these models for 1000 epochs with 10-fold cross validation. However, this empirical study in alignment with the CAL framework, is trained for 20 epochs with 5-fold cross validation.

\usetikzlibrary{patterns}
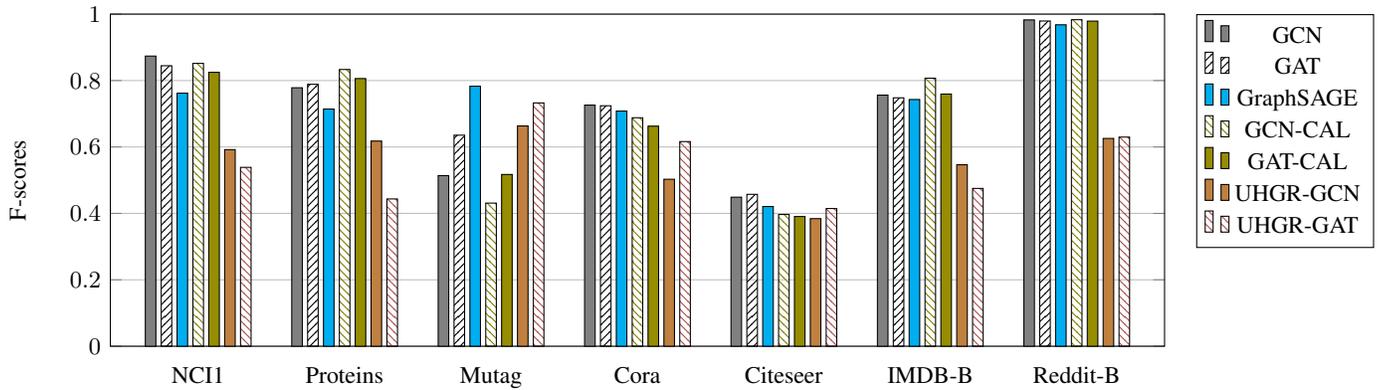
\begin{figure*}[h]
    \centering
    
    \begin{tikzpicture} [font=\small]  
    
        \begin{axis}[
            width  = 0.86*\textwidth,
            height = 6cm,
            major x tick style = transparent,
            ybar,
            bar width=4pt,
            ymajorgrids = true,
            ylabel = {F-scores},
            ymin=0, ymax = 1,
            symbolic x coords={NCI1, Proteins, Mutag, Cora, Citeseer, IMDB-B, Reddit-B},
            xtick = data,
            scaled y ticks = false,
            legend pos=outer north east,
        ]

            \addplot[style={black,fill=gray,mark=none}]
            coordinates { (NCI1, 0.8737)  (Proteins, 0.7783) (Mutag,  0.5138) (Cora, 0.7261) (Citeseer, 0.4488) (IMDB-B, 0.7562) (Reddit-B, 0.9826)};
            
            \addplot[style={black,fill=gray,pattern=north east lines, mark=none}]
            coordinates { (NCI1, 0.8446)  (Proteins, 0.7888) (Mutag, 0.6355) (Cora, 0.7241) (Citeseer,0.4574) (IMDB-B, 0.7480) (Reddit-B, 0.9793)};
            
            \addplot[style={black,fill=cyan,mark=none}]
            coordinates { (NCI1, 0.7621)  (Proteins,0.7142) (Mutag, 0.7830) (Cora, 0.7081) (Citeseer,0.4207) (IMDB-B, 0.7429) (Reddit-B, 0.9679)};
            
            \addplot[style={black,fill=lightgray,pattern=north west lines, pattern color=olive, mark=none}]
            coordinates { (NCI1, 0.8517)  (Proteins,0.8333) (Mutag,  0.4310) (Cora, 0.6880) (Citeseer,0.3967) (IMDB-B, 0.8071) (Reddit-B,0.9832)};
            
             \addplot[style={black,fill=olive,mark=none}]
            coordinates { (NCI1,0.8248)  (Proteins,0.8060) (Mutag,  0.5172) (Cora, 0.6627) (Citeseer,0.3906) (IMDB-B, 0.7593) (Reddit-B,0.9790)};

             \addplot[style={black,fill=brown,mark=none}]
            coordinates { (NCI1,0.5920)  (Proteins,0.6178) (Mutag, 0.6633) (Cora, 0.5028) (Citeseer,0.3842) (IMDB-B, 0.5465) (Reddit-B,0.6257)};

            \addplot[style={black,fill=lightgray,pattern=north west lines, pattern color=red!40!gray, mark=none}]
            coordinates { (NCI1, 0.5386)  (Proteins,0.4435) (Mutag, 0.7326) (Cora,0.6162) (Citeseer, 0.4146) (IMDB-B, 0.4752) (Reddit-B,0.6297)};

            \legend{GCN, GAT, GraphSAGE, GCN-CAL, GAT-CAL, UHGR-GCN, UHGR-GAT};
        \end{axis}
        
    \end{tikzpicture}
    
    \caption{F-scores for the key models on datasets}
    \label{fig:Fscore}
\end{figure*}

\usetikzlibrary{patterns}


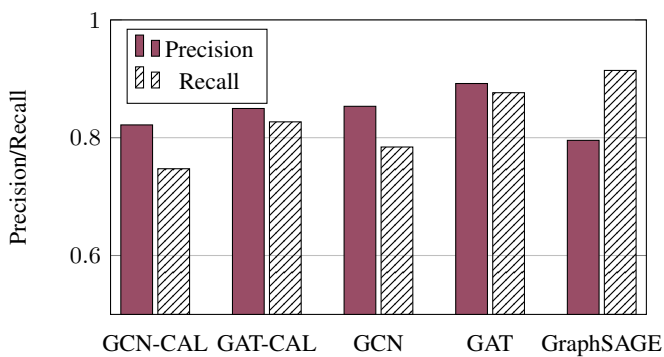
\begin{figure}[h]
    
    \begin{tikzpicture} [font=\small]  
    
        \begin{axis}[
            width  = 0.48*\textwidth,
            height = 5.5cm,
            major x tick style = transparent,
            ybar,
            bar width=12pt,
            ymajorgrids = true,
            ylabel = {Precision/Recall},
            ymin=0.5, ymax = 1,
            symbolic x coords={GCN-CAL, GAT-CAL, GCN, GAT, GraphSAGE},
            xtick = data,
            scaled y ticks = false,
            legend pos=north west,
        ]
            \addplot[style={black,fill=purple!40!gray,mark=none}]
                coordinates { (GCN-CAL, 0.8217) (GAT-CAL,0.8496) (GCN,0.8533)  (GAT,0.892) (GraphSAGE, 0.7955)};
    
            \addplot[style={black,fill=red,pattern=north east lines, mark=none}]
                coordinates { (GCN-CAL, 0.7473) (GAT-CAL,0.8269) (GCN,0.7842)  (GAT,0.8764) (GraphSAGE, 0.9142)}; 
    
            \legend{Precision,Recall};
        \end{axis}
    \end{tikzpicture}
    \caption{Precision-Recall Scores for the \textit{Mutag} dataset}
    \label{fig:PRMutag}
\end{figure}

\begin{figure}[h]
    \centering
    
    \begin{tikzpicture} [font=\small]  
    
        \begin{axis}[
            width  = 0.48*\textwidth,
            height = 5.5cm,
            major x tick style = transparent,
            ybar,
            bar width=12pt,
            ymajorgrids = true,
            ylabel = {Precision/Recall},
            ymin=0.5, ymax = 1,
            symbolic x coords={GCN-CAL, GAT-CAL, GCN, GAT, GraphSAGE},
            xtick = data,
            scaled y ticks = false,
            legend pos=north west,
        ]
            \addplot[style={black,fill=purple!40!gray,mark=none}]
                coordinates { (GCN-CAL, 0.8133) (GAT-CAL,0.8111) (GCN,0.7956)  (GAT,0.7697) (GraphSAGE, 0.8344)};
    
            \addplot[style={black,fill=red,pattern=north east lines, mark=none}]
                coordinates { (GCN-CAL, 0.7179) (GAT-CAL,0.7657) (GCN,0.7425)  (GAT,0.7587) (GraphSAGE, 0.7814)}; 
    
            \legend{Precision,Recall};
        \end{axis}
    \end{tikzpicture}
    \caption{Precision-Recall Scores for the \textit{Cora} dataset}
    \label{fig:PRCora}
\end{figure}
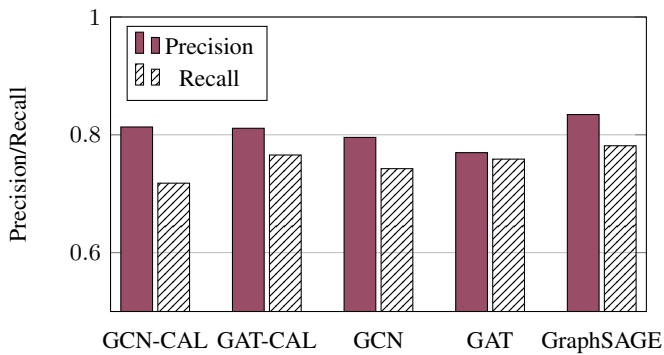

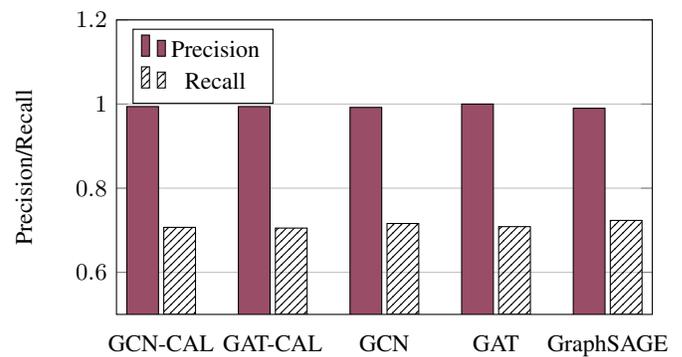
\begin{figure}[h]
    
    \begin{tikzpicture} [font=\small]  
    
        \begin{axis}[
            width  = 0.48*\textwidth,
            height = 5.5cm,
            major x tick style = transparent,
            ybar,
            bar width=12pt,
            ymajorgrids = true,
            ylabel = {Precision/Recall},
            ymin=0.5, ymax = 1.2,
            symbolic x coords={GCN-CAL, GAT-CAL, GCN, GAT, GraphSAGE},
            xtick = data,
            scaled y ticks = false,
            legend pos=north west,
        ]
            \addplot[style={black,fill=purple!40!gray,mark=none}]
                coordinates { (GCN-CAL, 0.994) (GAT-CAL,0.994) (GCN,0.992)  (GAT,1) (GraphSAGE, 0.99)};
    
            \addplot[style={black,fill=red,pattern=north east lines, mark=none}]
                coordinates { (GCN-CAL, 0.7069) (GAT-CAL,0.7052) (GCN,0.7161)  (GAT,0.7086) (GraphSAGE, 0.7234)}; 
    
            \legend{Precision,Recall};
        \end{axis}
    \end{tikzpicture}
    \caption{Precision-Recall Scores for the \textit{IMDB-Binary} dataset}
    \label{fig:PRIMDBB}
\end{figure}

 \begin{table}[hbt!]
 \fontsize{8pt}{10pt}\selectfont
\caption{Test Accuracy (\%) of classification tasks}
 \centering
  \begin{tabular}
  {p{2.05cm} p{0.4cm} p{0.52cm} p{0.5cm} p{0.4cm} p{0.6cm} p{0.5cm} p{0.5cm}}
 
  \hline
    Model  & NCI1   & Proteins  & Mutag & Cora & Citeseer & IMDB-B & Reddit-B  \\
    
  \hline

    GCN \cite{kipf2016semi} & 80.68 & 75.47 & 84.10 & 79.74 & 65.30 & 73.60 & 91.25    \\
    GAT \cite{velivckovic2017graph} & 79.54  & 75.02 & 89.42  & 83.03 & 68.41 & 72.00 & 90.85  \\
    GIN \cite{xu2018powerful}  & 80.44 & 75.29 & 85.14  & 80.82 & 61.80 & 72.50 & 88.75      \\
    GraphSAGE \cite{hamilton2017inductive}  & 74.21 & 75.83 & 83.49  & 76.55 & 61.66 & 73.00 & 77.20  \\
    GCN-CAL \cite{sui2022causal}  & 81.07 & 75.29 & 78.68 & 80.21 & 61.74 & 72.70 & 91.15 \\
    GAT-CAL \cite{sui2022causal}  & 80.49 & 75.56 & 86.26 & 83.16 & 65.61 & 73.20 & 91.30 \\
    GIN-CAL \cite{sui2022causal}  & 80.36 & 73.76 & 71.86 & 80.53 & 59.97 & 72.60 & 89.65 \\
    UHGR-GAT \cite{ding2020unsupervised} & 59.80 & 70.09 & 77.77 & 67.55 & 53.30 & 55.00 & 64.66 \\
    UHGR-GCN \cite{ding2020unsupervised} & 61.07 & 67.92 & 74.11 & 62.65 & 50.25 & 56.20 & 64.77 \\
          \hline

      \end{tabular}
  \label{tab:tablStudyM}
\end{table}

\subsubsection*{Evaluation}
Evaluation was performed on two tasks viz. node classification using citation datasets and graph classification using the bio-chemical and social datasets. The evaluation metrics used were classification accuracy, Precision/Recall scores and F1-Scores. Accuracy is the percentage of accurately predicted data. Precision and recall are evaluation metrics that provide insights into a model's accuracy in making positive predictions and the comprehensiveness of these positive predictions, respectively. The F-Score consolidates the precision and recall measures into a single metric.

\subsection{Experimental Results}

The test accuracy results of these studies are summarised in Tab.~\ref{tab:tablStudyM}. The results for the citation datasets are for the node classification task and the remaining results are for the graph classification task. 
The precision/recall results for GCN-CAL, GAT-CAL, GCN, GAT and GraphSAGE are plotted for one dataset each from the three dataset categories in Fig.~\ref{fig:PRMutag} (Mutag dataset), Fig.~\ref{fig:PRCora} (Cora dataset) and Fig.~\ref{fig:PRIMDBB} (IMDB-B dataset). The F1-scores for GCN, GAT, GraphSAGE, GCN-CAL, GAT-CAL, UHGR-GCN and UHGR-GAT models on all the datasets are shown in Fig.~\ref{fig:Fscore}.

\begin{figure}[h]
    \centering
    \includegraphics[width=\columnwidth]{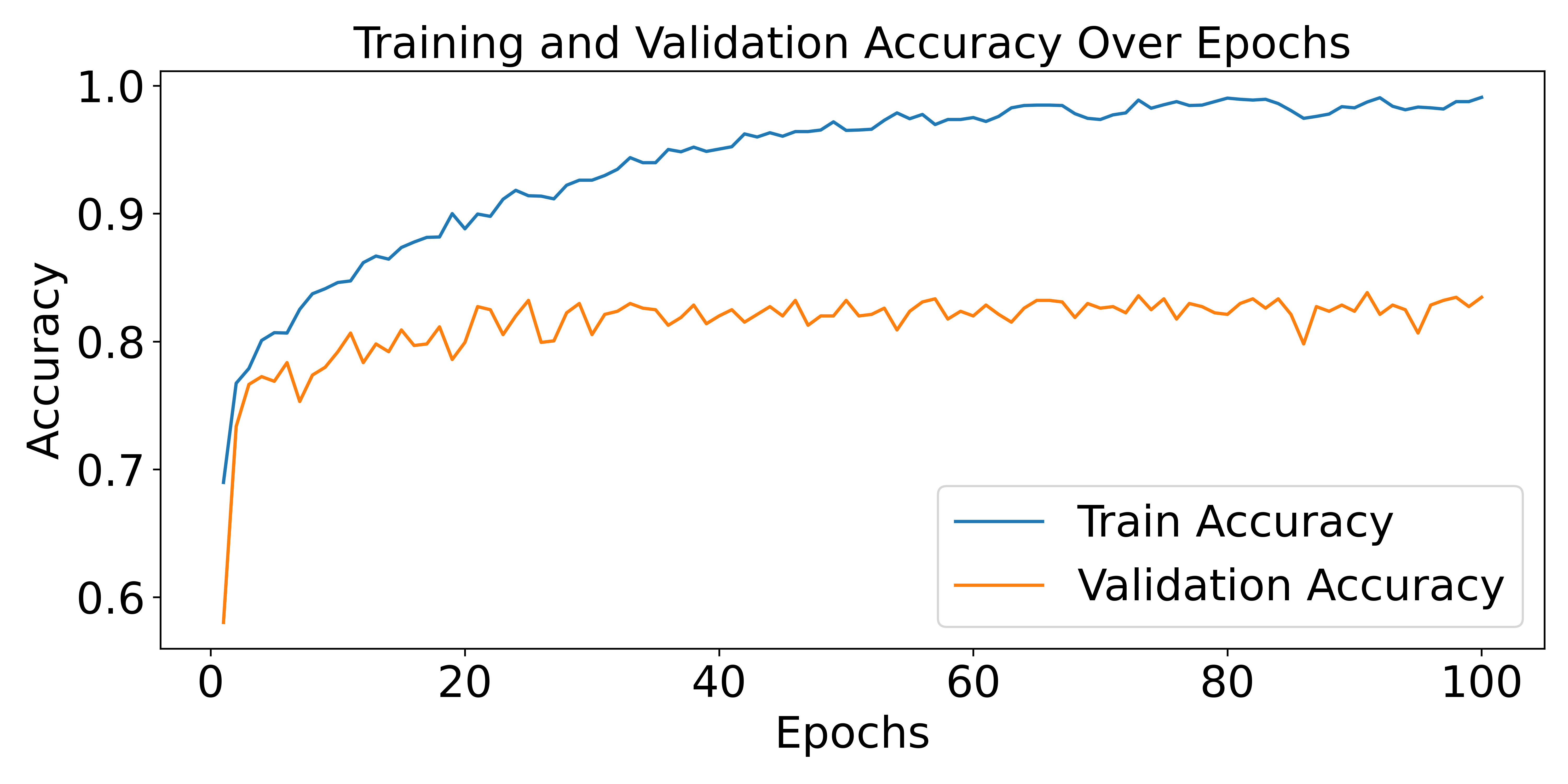}
    \caption{Epochs vs. Accuracy plot for \textit{GCN-CAL} model (\textit{NCI1} dataset)}
    \label{fig:cal_gcn_nci1}
\end{figure}

\begin{figure}[h]
    \centering
    \includegraphics[width=\columnwidth]{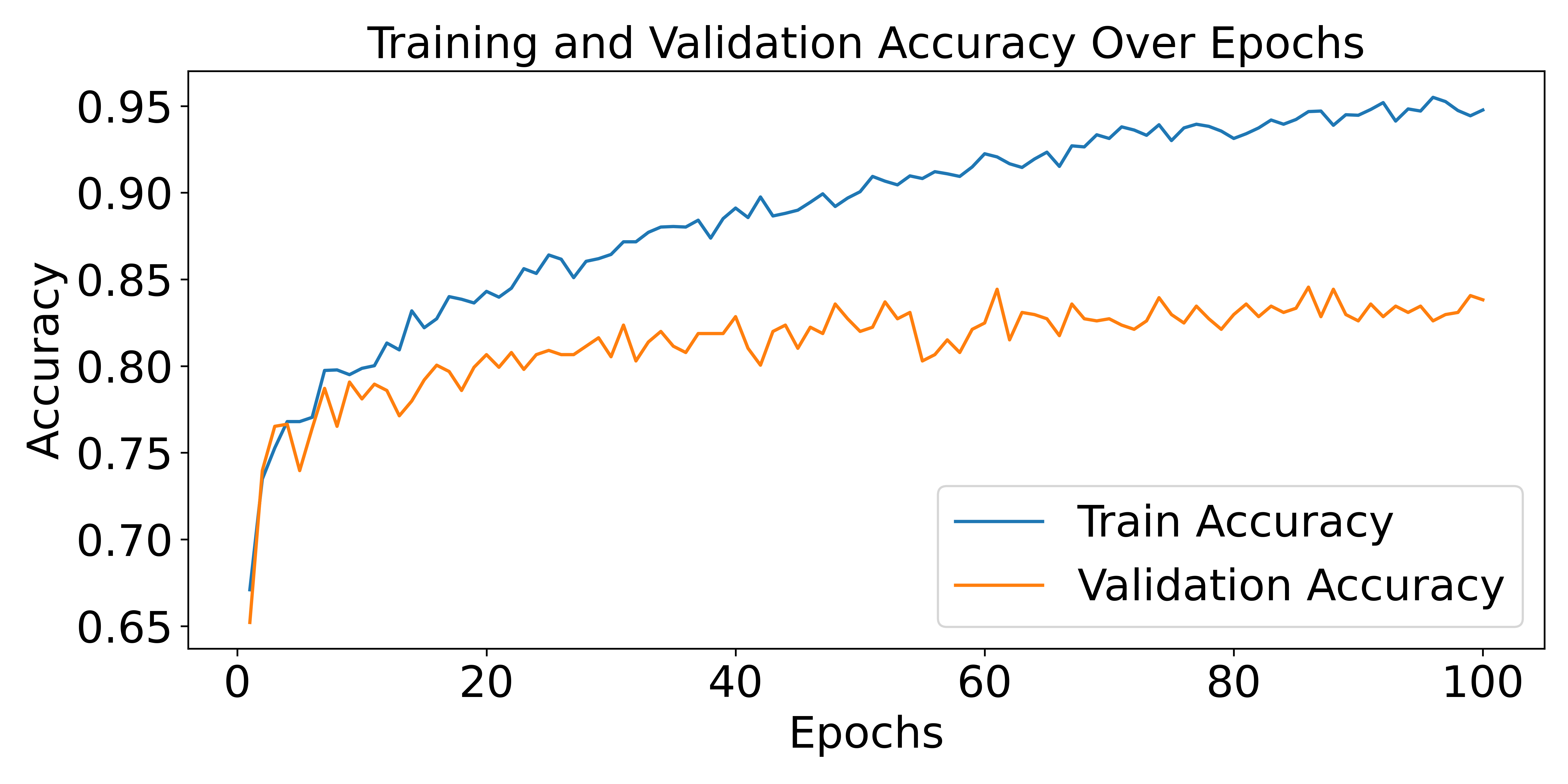}
    \caption{Epochs vs. Accuracy plot for \textit{GAT-CAL} model (\textit{NCI1} dataset)}
    \label{fig:cal_gat_nci1}
\end{figure}

\begin{figure}[h]
    \centering
    \includegraphics[width=\columnwidth]{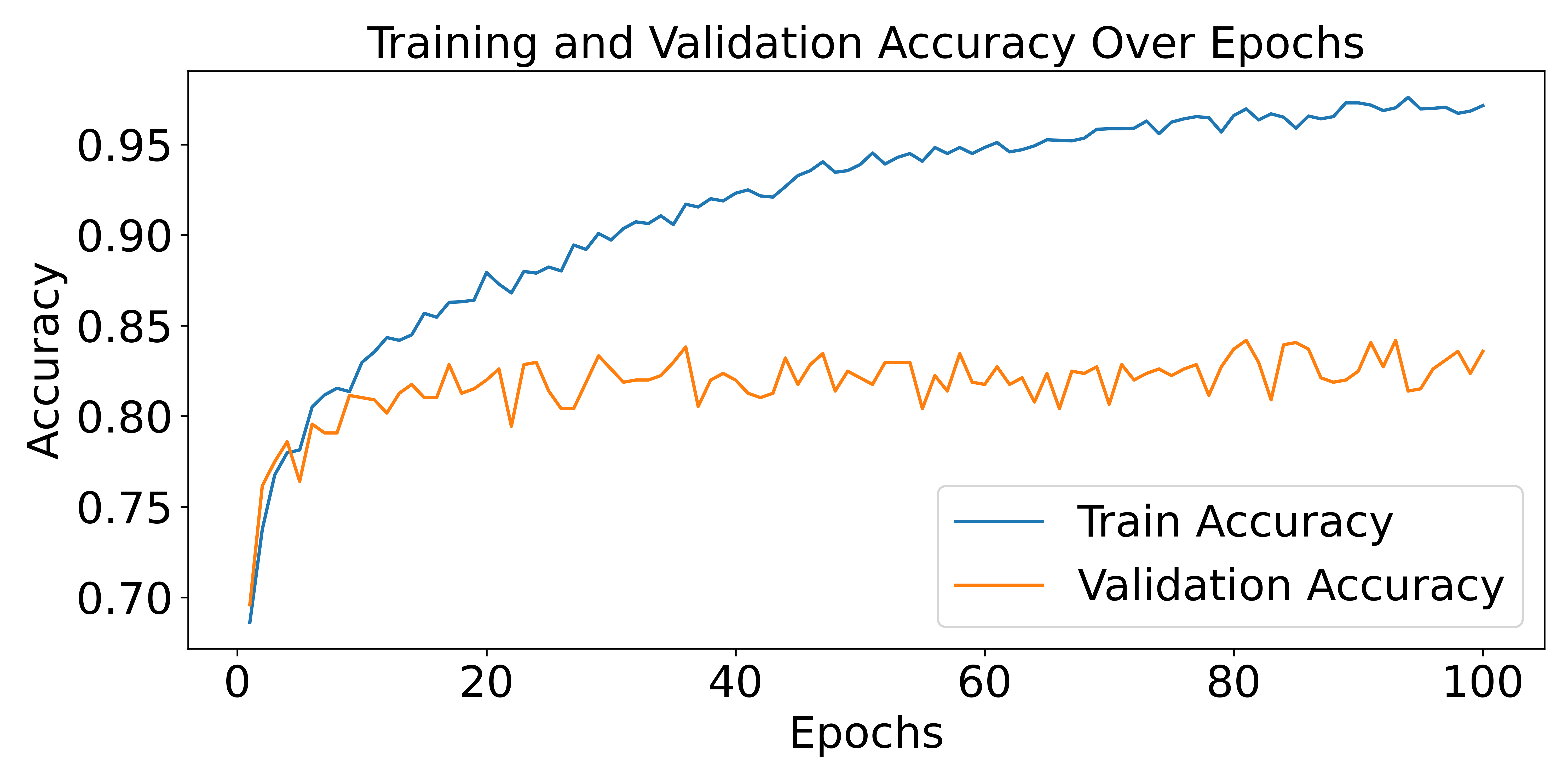}
    \caption{Epochs vs. Accuracy plot for \textit{GCN} model (\textit{NCI1} dataset)}
    \label{fig:gcn_nci1}
\end{figure}

\begin{figure}[h]
    \centering
    \includegraphics[width=\columnwidth]{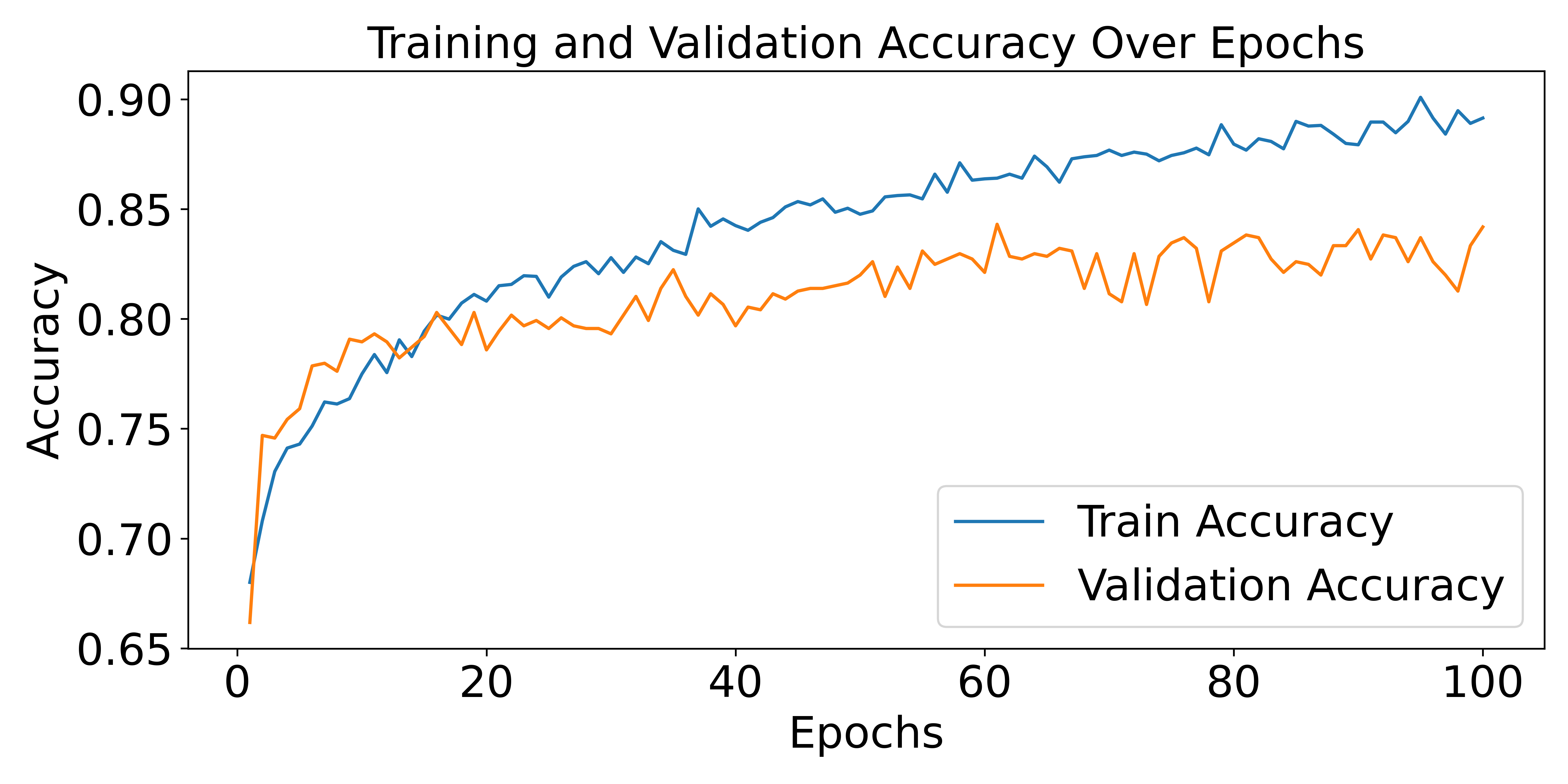}
    \caption{Epochs vs. Accuracy plot for \textit{GAT} model (\textit{NCI1} dataset)}
    \label{fig:gat_nci1}
\end{figure}

\begin{figure}[h]
    \centering
    \includegraphics[width=\columnwidth]{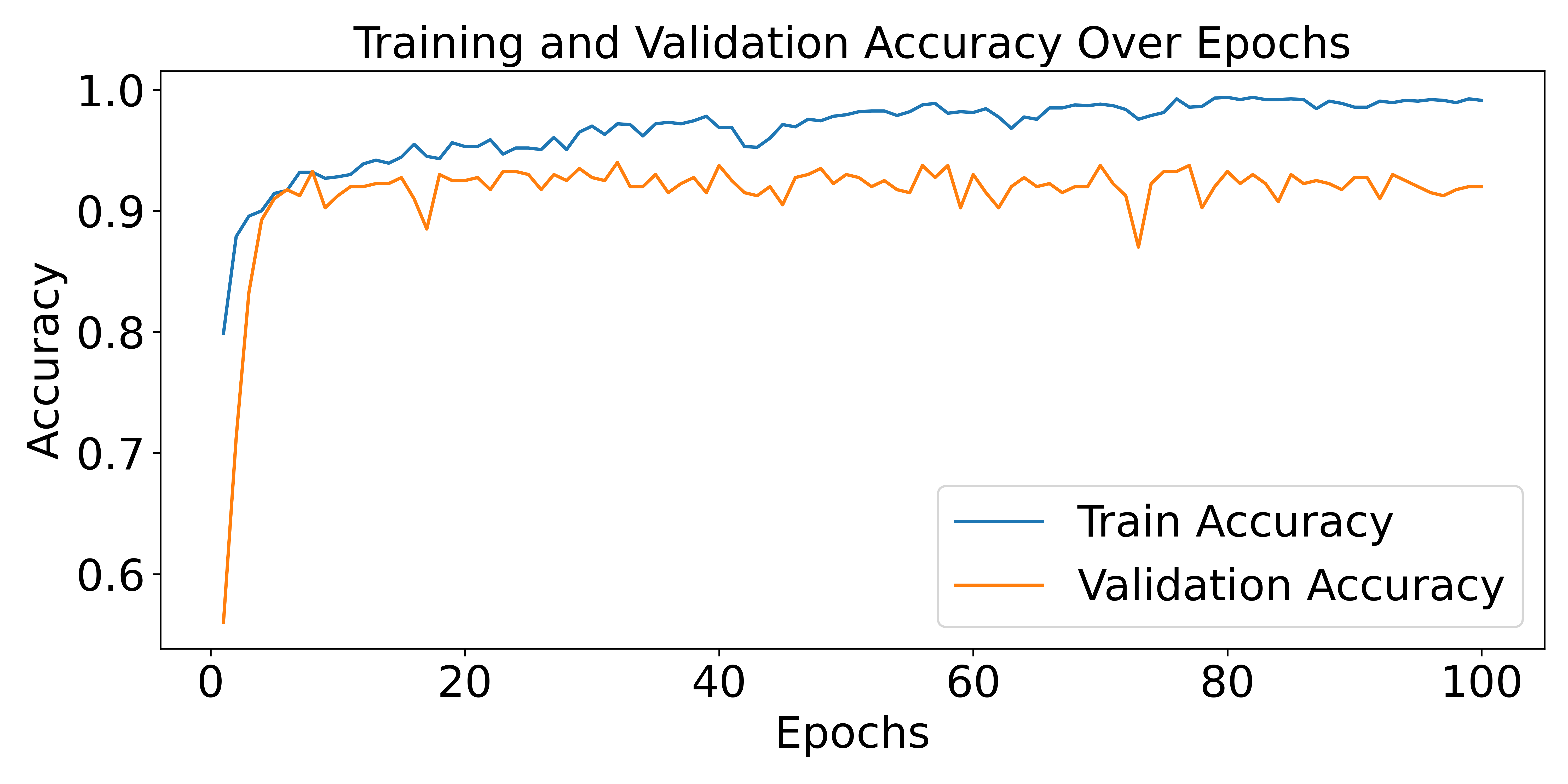}
    \caption{Epochs vs. Accuracy plot for \textit{GCN-CAL} model (\textit{Reddit-B} dataset)}
    \label{fig:cal_gcn_RedditB}
\end{figure}

\begin{figure}[h]
    \centering
    \includegraphics[width=\columnwidth]{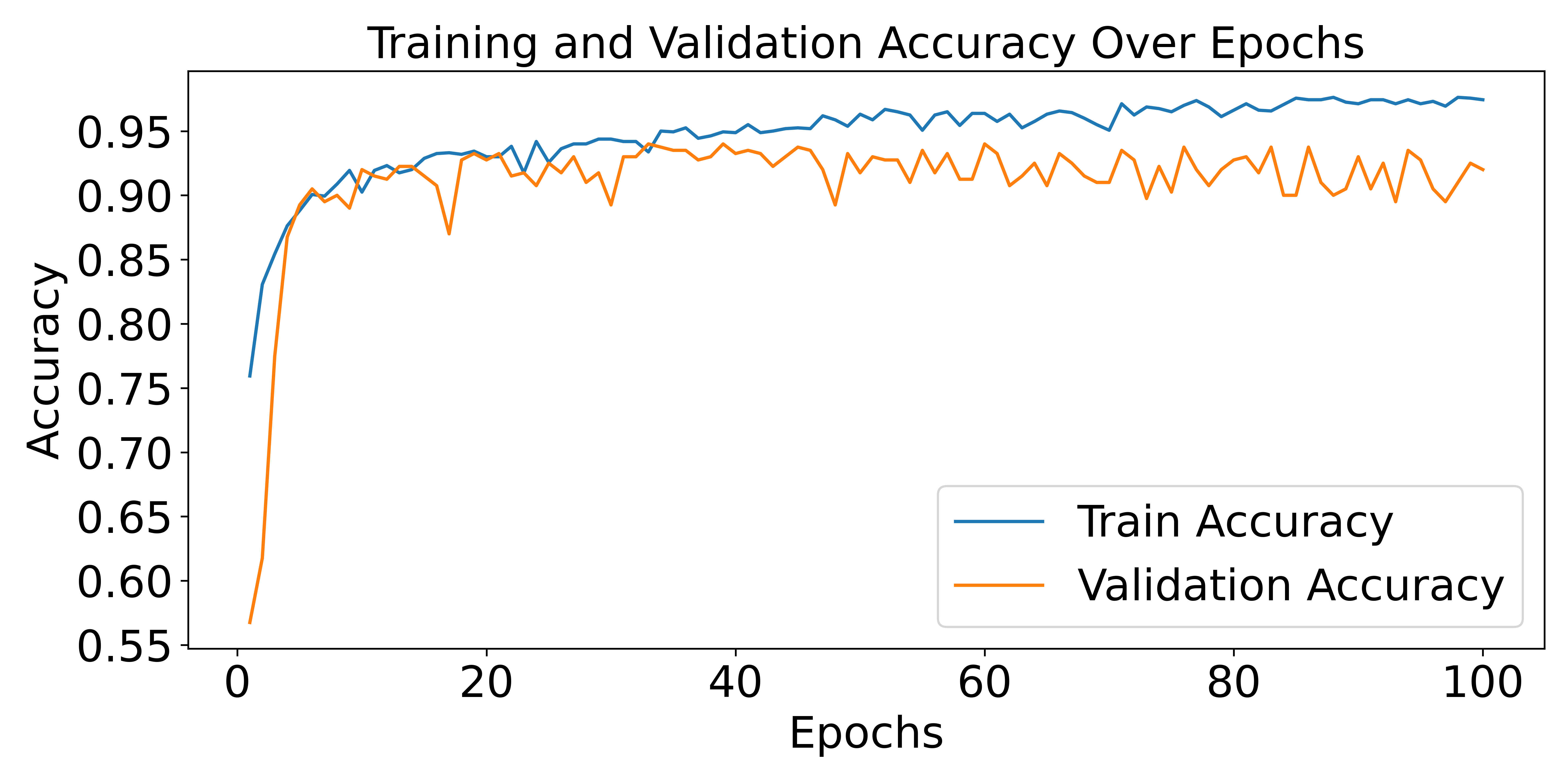}
    \caption{Epochs vs. Accuracy plot for \textit{GAT-CAL} model (\textit{Reddit-B} dataset)}
    \label{fig:cal_gat_RedditB}
\end{figure}

\begin{figure}[h]
    \centering
    \includegraphics[width=\columnwidth]{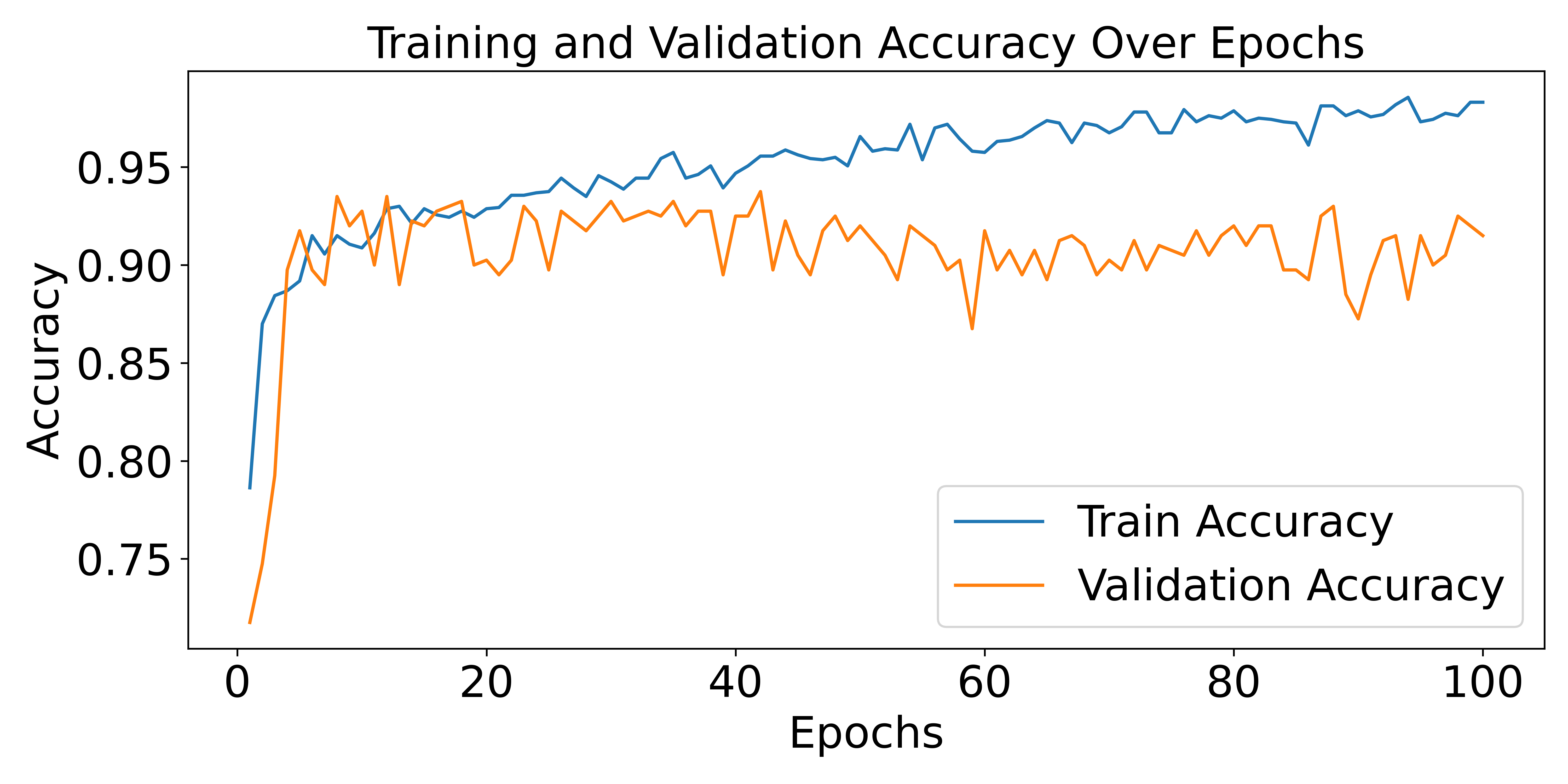}
    \caption{Epochs vs. Accuracy plot for \textit{GCN} model (\textit{Reddit-B} dataset)}
    \label{fig:gcn_RedditB}
\end{figure}

\begin{figure}[h]
    \centering
    \includegraphics[width=\columnwidth]{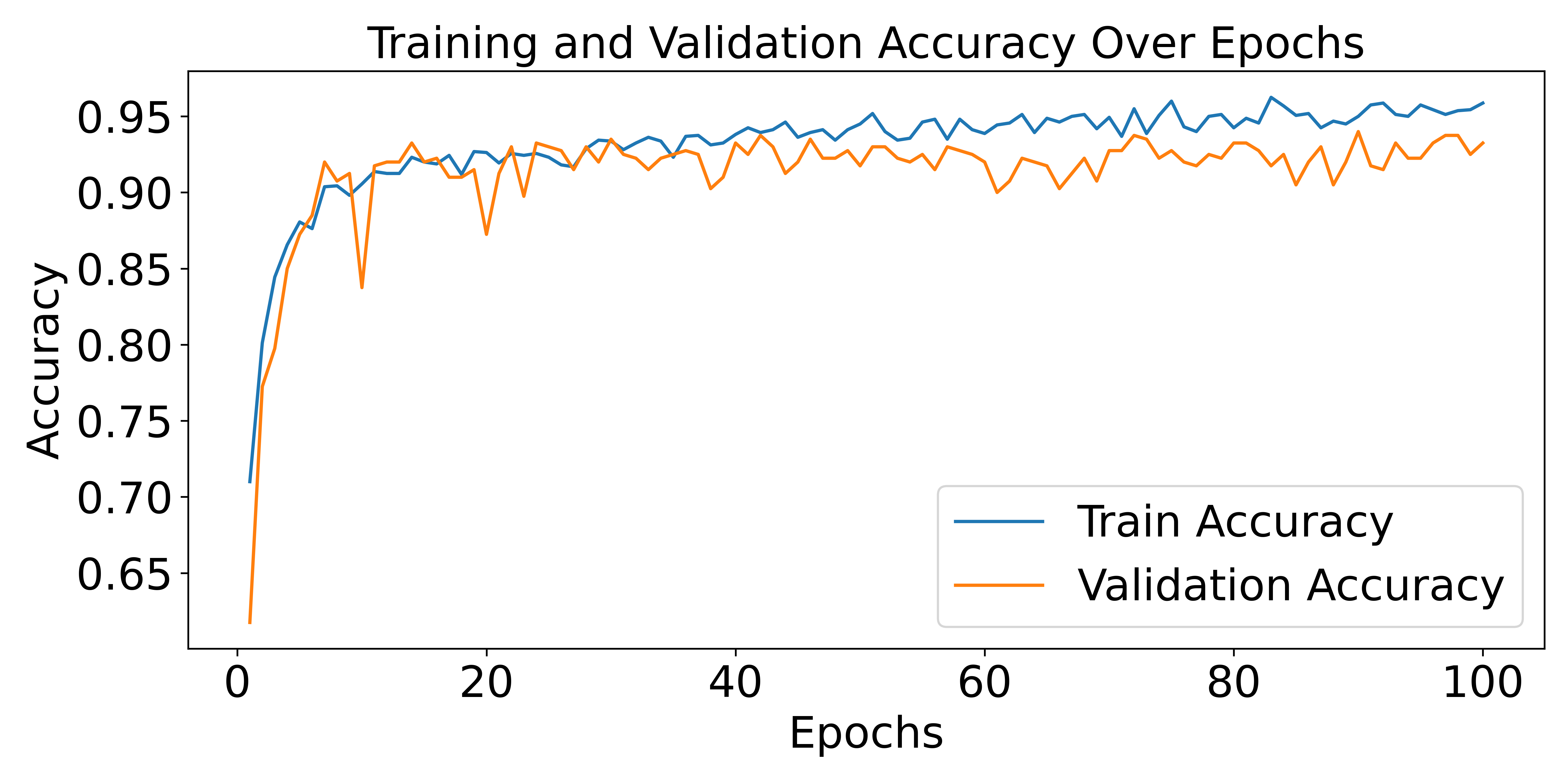}
    \caption{Epochs vs. Accuracy plot for \textit{GAT} model (\textit{Reddit-B} dataset)}
    \label{fig:gat_RedditB}
\end{figure}

Further to this, the training and validation accuracy over 100 epochs are plotted and the model performance is studied. The performance of the GCN-CAL framework on the NCI1 and the Reddit-B datasets are illustrated in Fig.~\ref{fig:cal_gcn_nci1} and Fig.~\ref{fig:cal_gcn_RedditB} respectively. Similarly, the performance of the GAT-CAL framework on the NCI1 and the Reddit-B datasets are illustrated in Fig.~\ref{fig:cal_gat_nci1} and Fig.~\ref{fig:cal_gat_RedditB} respectively. The performance of the GCN framework on the NCI1 and the Reddit-B datasets are illustrated in Fig.~\ref{fig:gcn_nci1} and Fig.~\ref{fig:gcn_RedditB} respectively. Similarly, the performance of the GAT framework on the NCI1 and the Reddit-B datasets are illustrated in Fig.~\ref{fig:gat_nci1} and Fig.~\ref{fig:gat_RedditB} respectively.

\subsubsection*{\textbf{Discussion}}

On analysing the classification performance results in terms of accuracy scores, a higher performance of the GAT architecture is observed among the base models for most of the datasets. The classification framework augmented with causal elements, CAL \cite{sui2022causal} demonstrated the model's efficiency when causality is incorporated into the framework. This was specifically observed with the GAT-CAL model, for which the performance was higher compared to other architectures, except for the NCI1 dataset. For the UHGR models \cite{ding2020unsupervised}, experimentation on the seven datasets demonstrated significant decline in performance for the social network datasets and the NCI1 dataset, when compared to the other models as shown in Tab.~\ref{tab:tablStudyM}. The precision and recall scores are mostly evenly distributed for the Mutag and Cora datasets. However, the scores for the IMDB-B dataset exhibit a substantial difference.   

Upon analyzing the F1-scores, it's evident that performance varies across algorithms on the Mutag dataset. Apart from this, all the models demonstrate effective generalization across all datasets, with the exception of the UHGR model on NCI1 and the social network datasets. The UHGR-GAT model exhibits a decline in performance on the protein dataset as well. Overall, it is observed from the F1-scores that GCN and GAT performs better than the other models on node classification tasks. For graph classification tasks, the CAL framework presents a promising prospect and is mostly generalizable across datasets, with a dip in performance with the Mutag dataset. 
The performance of the GraphSAGE architecture has been notable on the Mutag dataset, whereas all other models exhibit lower performance for this dataset. While the CAL framework has generally shown comparable performance with other models, except on the Mutag dataset, the CAL framework's failure to achieve notably high results suggests a limitation in causal modeling. Similarly, though the UHGR framework has demonstrated results on par with the other models for the Mutag and citation datasets, this model has also fallen short of achieving exception results. This implies that MI is a significant contributor to the classification performance. However, additional enhancements are required to improve the model further. 

Here, we discuss some of the advantages and disadvantages of these models.

\begin{itemize}
    \item GCN \cite{kipf2016semi} has higher computational efficiency and is particularly suitable when local neighbourhood information is relevant for the problem. This characteristic is notably reflected in its high F-score values for node classification.

    \item GAT \cite{velivckovic2017graph} is better suited for inductive settings and excels at capturing intricate relationships through its attention mechanism. However, these advantages come with a few drawbacks, including the computational complexity and hyperparameter sensitivity of the GAT architecture. The sensitivity to L2 regularization and learning rates are particularly evident for the its performance on the Mutag and citation datasets respectively.

    \item GraphSAGE \cite{hamilton2017inductive} is a framework that is suitable for large-scale graphs. A limitation of this architecture is that it relies on sampling a fixed-size neighbourhood for each node, and hence cannot capture distant information beyond this boundary. It is noted that GraphSAGE demonstrates greater robustness in capturing social networks compared to molecular graphs, which could be attributed to its limited capacity to capture complex molecular structures.
    
    \item GIN \cite{xu2018powerful} is more adept at capturing global graph structures as compared to other architectures. However, GIN does not incorporate node features during aggregation, leading to loss of significant node information. On the other hand, GCN offers a more comprehensive node representation by leveraging node features in the aggregation function. This is evident in the consistently strong performance of GIN in graph classification tasks, as indicated by the high F1-scores and accuracy rates.

    \item The CAL framework adopts a causal classification approach utilizing the attention mechanism. While this approach leverages weighted information to infer causality and capture dependencies among causal features, handling long-range causality may pose challenges for this model. With the exception of Mutag, all three CAL models demonstrate strong performance across all datasets. However, the observed performance improvement is not substantial when compared to the baseline GCN, GAT, and GIN models. Hence, the outcomes of our study do not conclusively establish the effectiveness of the CAL framework on classification performance.

    \item The UHGR framework utilized  mutual information maximization (MIM) for classification tasks. MIM has the capability to learn meaningful representations by capturing both local and global data patterns. The UHGR model demonstrated positive performance outcomes in node classification tasks, emphasizing the significance of mutual information in understanding dependencies among individual nodes. However, the graph classification performance results do not indicate significant improvements for base models augmented with MIM. It is also essential to recognize that MIM does not inherently produce causal representations.
   
\end{itemize}
    In summary, it is important to emphasize that these five frameworks do not inherently capture causal elements, necessitating additional steps for the integration of causality in classification tasks.

\subsubsection{Sensitivity Analysis} 

A sensitivity study has been performed for all the models based on learning rates and L2 regularization or weight decay. The learning rate (LR) and weight decay (WD) are 1e-3 and 0 in the original studies. The results for varying learning rates with WD=0 are shown in Tab.~\ref{tab:tablehyper}. The results for varying weight decay rates for LR=1e-3 are shown in Tab.~\ref{tab:tablehyperWD}. The results of the analysis are summarised as follows:

\begin{itemize}
    \item It is evident from the experimental findings that different learning rates result in diverse performances for all models across the entirety of datasets. A pronounced variation in F-scores is noticeable for all models, except the UHGR models, across the Mutag, Cora, and Citeseer datasets. This suggests that the impact of the learning rate is predominantly influenced by dataset characteristics, as opposed to the model architecture.

  \item The impact of varying the weight decay rate is particularly evident in the case of the two UHGR models, as all figures exhibit noticeable differences, with a potential exception for UHGR-GCN on the Cora dataset. The CAL framework displays variances primarily on the citation datasets, suggesting a potential sensitivity of the model to node classification tasks concerning WD. The impact of varying WD on other models is marginal, except for GCN and GAT on the Mutag dataset, where noticeable effects are observed. This suggests the dataset's sensitivity to the regularization mechanism.
\end{itemize}

\renewcommand{\arraystretch}{2}
\begin{table*}[hbt!]
\fontsize{7.6pt}{8pt}\selectfont
\caption{Classification F-Scores  with time (secs) for varying Learning Rates \label{tab:tablehyper}}
  \centering
  \setlength\extrarowheight{-2pt}
   \begin{tabular}  {p{1.8cm} c c c c c c c c c c c c c c c}

  \hline

  \textbf{Model} & LR &  \multicolumn{2}{c}  {\parbox{1.5cm}{\centering NCI1 \\        }} & \multicolumn{2}{c} {\parbox{1.5cm}{\centering Proteins \\        }}  & \multicolumn{2}{c} {\parbox{1.5cm}{\centering Mutag \\        }} & \multicolumn{2}{c} {\parbox{1.5cm}{\centering Cora \\        }}  & \multicolumn{2}{c} {\parbox{1.5cm}{\centering Citeseer \\        }}  & \multicolumn{2}{c} {\parbox{1.5cm}{\centering IMDB-B \\        }} & \multicolumn{2}{c} {\parbox{1.5cm}{\centering REDDIT-B \\        }}  \\

  & & time & f-score & time & f-score & time & f-score & time & f-score & time & f-score & time & f-score & time & f-score \\
  
\hline

{\multirow{1}{*}{\textbf{GCN \cite{kipf2016semi}}}} & 1e-2 & 41.8 & 0.8117 & 17.9 & 0.7141 & 7.9 & 0.7829 & 51.7 & 0.6814 & 43.1 & 0.5067 & 14.2 & 0.7904 & 51.1 & 0.9616 \\

 & 1e-3 & 45.1 & 0.8737 & 17.4 & 0.7783 & 7.6 & 0.5138 & 51.4 & 0.7261 & 49.7 & 0.4488 & 14.8 & 0.7562 & 54.6 & 0.9826 \\
  & 1e-4 & 44.6 & 0.8408 & 17.7 & 0.8289 & 7.5 & 0.3875 & 53.0 & 0.4598 & 48.6 & 0.3476 & 14.8 & 0.8246 & 56.3 & 0.9905 \\
   & 1e-5 & 43.7 & 0.7228 & 17.7 & 0.7026 & 7.6 & 0.3770 & 51.4 & 0.2189 & 53.5 & 0.1779 & 14.8 & 0.7583 & 47.4 & 0.9628 \\
    \hline

 {\multirow{1}{*}{\textbf{GAT \cite{velivckovic2017graph}}}} & 1e-2 & 48.0 & 0.8181 & 19.6 & 0.7205 & 8.1 & 0.7835 & 64.8 & 0.7100 & 48.3 & 0.5802 & 14.6 & 0.7676 & 64.6 & 0.9508 \\

 & 1e-3 & 47.3 & 0.8446 & 19.1 & 0.7888 & 8.2 & 0.6355 & 54.9 & 0.7241 & 53.4 & 0.4574 & 15.4 & 0.7480 & 65.8 & 0.9793 \\
  & 1e-4 & 49.1  & 0.7837 & 18.9 & 0.8544 & 8.2 & 0.5316 & 54.0 & 0.4488 & 52.0 & 0.2538 & 15.7 & 0.7735 & 70.5 & 0.9967  \\
   & 1e-5 & 47.6 & 0.7018 & 19.3 & 0.8247 & 8.1 & 0.4763 & 53.8 & 0.2430 & 52.08 & 0.1511 & 15.7 & 0.7523 & 67.4 & 0.9241 \\
    \hline

   {\multirow{1}{*}{\textbf{GIN \cite{xu2018powerful}}}} & 1e-2 & 39.7 & 0.7246 & 16.2 & 0.6833 & 7.2 & 0.6376 & 33.3 & 0.5789 & 30.3 & 0.3058 & 49.1 & 0.6456 & 49.5 & 0.8820 \\

 & 1e-3 & 39.9 & 0.8507 & 16.0 & 0.7800 & 7.1 & 0.6929 & 30.8 & 0.6953 & 30.6 & 0.4258 & 18.1 & 0.6609 & 49.2 & 0.9947  \\
  & 1e-4 & 40.3 & 0.8486 & 16.2 & 0.8629 & 7.1 & 0.4843 & 36.9 & 0.4234 & 30.7 & 0.3141 & 14.5 & 0.8482 & 51.3 & 0.9970 \\
   & 1e-5 & 39.0 & 0.7891 & 16.2 & 0.8754 & 7.2 & 0.4112 & 43.0 & 0.2199 & 30.2 & 0.1617 & 12.8 & 0.7468 & 53.9 & 0.9864 \\
    \hline

     {\multirow{1}{*}{\textbf{GraphSAGE \cite{hamilton2017inductive}}}} & 1e-2 & 42.3 & 0.7610 & 16.5 & 0.7631 & 7.8 & 0.8011 & 37.8 &  0.6579 & 37.3 & 0.4661 & 19.3 & 0.7982 & 49.9 & 0.9615 \\

 & 1e-3 & 43.8 & 0.7621 & 17.3 & 0.7142 & 7.9 & 0.7830 & 37.9 & 0.7081 & 37.0 & 0.4207 & 18.5 & 0.7429 & 71.7 & 0.9679 \\
  & 1e-4 & 45.1 & 0.7618 & 16.8 & 0.7723 & 7.8 & 0.6131 & 39.1 & 0.3051 & 35.4 & 0.2352 & 19.7 & 0.6979 & 48.5 & 0.9867 \\
   & 1e-5 & 43.4 & 0.8637 & 17.5 & 0.7790 & 7.9 & 0.4720 & 38.2 & 0.2208 & 36.5 & 0.1905 & 30.9 & 0.6585 & 61.5 & 0.8499 \\
    \hline

     {\multirow{1}{*}{\textbf{GCN-CAL \cite{sui2022causal}}}} & 1e-2 & 57.4 & 0.8331 & 21.7 & 0.7640 & 9.1 & 0.7453 & 72.2 & 0.7327 & 70.6 & 0.4944 & 18.9 & 0.8069 & 105.4 & 0.9688 \\

 & 1e-3 & 55.9 & 0.8517 & 21.4 & 0.8333 & 9.5 & 0.4310 & 74.7 & 0.6880 & 76.5 & 0.3967 & 19.1 & 0.8071 & 104.9 & 0.9832 \\
  & 1e-4 & 54.4 & 0.7732 & 21.5 & 0.8803 & 9.5 & 0.5061 & 71.8 & 0.3337 & 74.8 & 0.2294 & 18.9 & 0.8247 & 97.8 & 0.9980 \\
   & 1e-5 & 59.2 & 0.6762 & 21.7 & 0.7096 & 9.4 & 0.4896 & 74.2 & 0.1955 & 76.0 & 0.1313 & 19.4 &  0.6995 & 97.0 & 0.8703 \\
    \hline

     {\multirow{1}{*}{\textbf{GAT-CAL \cite{sui2022causal}}}} & 1e-2 & 56.9 & 0.8299 & 22.6 & 0.7380 & 9.6 & 0.7560 & 79.9 & 0.7364 & 71.1 & 0.5182 & 20.6 & 0.8054 & 115.9 & 0.9453 \\

 & 1e-3 & 64.0  & 0.8248 & 22.3 & 0.8060 & 10.1 & 0.5172 & 79.0 & 0.6627 & 70.3 & 0.3906 & 20.5 & 0.7593 & 120.7 & 0.9790 \\
  & 1e-4 & 60.9 & 0.7261 & 22.2 & 0.8976 & 10.3 & 0.3713 & 79.7 & 0.2726 & 71.6 & 0.1856 & 20.1 & 0.7396 & 124.5 & 0.9912 \\
   & 1e-5 & 57.1 & 0.6790 & 22.8 & 0.8573 & 10.0 & 0.4282 & 91.9 & 0.2356 & 70.5 & 0.1603 & 19.3 & 0.6120 & 127.8 &  0.9819 \\
    \hline

     {\multirow{1}{*}{\textbf{GIN-CAL \cite{sui2022causal}}}} & 1e-2 & 55.6 & 0.7389 & 20.9 & 0.7675 & 8.8 & 0.6109 & 56.4 & 0.3470 & 59.3 & 0.2837 & 18.8 & 0.6336 & 99.3 & 0.9002 \\

 & 1e-3 & 50.0  & 0.8130 & 20.1 & 0.8492 & 8.9 & 0.2664 & 62.6 & 0.6174 & 61.9 & 0.3629 & 18.5 & 0.7051 & 100.0 & 0.9875 \\
  & 1e-4 & 49.3 & 0.7713 & 20.1 & 0.8993 & 9.1 & 0.2846 & 58.9 & 0.3372 & 53.9 & 0.2058 & 17.9 & 0.7739 & 99.3 & 0.9996 \\
   & 1e-5 & 53.8  & 0.7305  & 20.1 & 0.8769 & 9.1 & 0.3710 & 57.1 & 0.2765 & 57.4 & 0.1353 & 17.4 & 0.6832 & 101.9 & 0.9571 \\
    \hline

     {\multirow{1}{*}{\textbf{UHGR-GAT \cite{ding2020unsupervised}}}} & 1e-2 & 1693 & 0.5386 & 493.4 & 0.6277 & 96.2 & 0.7425 & 8.7 & 0.5864 & 9.5 & 0.4358 & 418.0 & 0.4653 & 798.2 & 0.6257 \\

 & 1e-3 & 1701 & 0.5386 & 449.6 & 0.4435 & 95.1 & 0.7326 & 8.7 & 0.6162 & 9.5 & 0.4146 & 416.8 & 0.4752 & 763.8 & 0.6297 \\
  & 1e-4 & 1743 & 0.4851 & 449.4 & 0.5603 & 96.7 & 0.7168 & 8.5 & 0.5926 & 9.3 & 0.3830 & 421.4 & 0.5524 & 751.5 & 0.6099 \\
   & 1e-5 & 1884 & 0.5702 & 453.6 & 0.6336 & 96.2 & 0.7287 & 8.9 & 0.5842 & 9.4 & 0.4624 & 412.6 & 0.4831 & 750.8 & 0.6376\\
    \hline
     {\multirow{1}{*}{\textbf{UHGR-GCN \cite{ding2020unsupervised}}}} & 1e-2 & 1672 & 0.5425 & 458.1 & 0.6633 & 94.6 & 0.7524 & 8.5 & 0.5224 & 9.0 & 0.4310 & 420.9 & 0.4910 & 758.9 & 0.6415 \\

 & 1e-3 & 1693 & 0.5920 & 456.8 & 0.6178 & 92.1 & 0.6633 & 8.1 & 0.5028 & 8.9 & 0.3842 & 417 & 0.5465 & 765.1 & 0.6257 \\
  & 1e-4 & 1764 & 0.5603 & 474.2 & 0.6673 & 94.8 & 0.7425 & 8.4 & 0.5308 & 9.5 & 0.4086 & 432.4 & 0.4633 & 746.9 & 0.5663 \\
   & 1e-5 & 1749 & 0.6198 & 466.6 &  0.6752 & 93.1 & 0.7940 & 8.4 & 0.5192 & 9.4 &  0.3902 & 421.0 & 0.5168 & 742.7 & 0.5782 \\
    \hline

  \end{tabular}
  \label{tab:tablehyper}
\end{table*}


\renewcommand{\arraystretch}{2}
\begin{table*}[hbt!]
\fontsize{7.6pt}{8pt}\selectfont
\caption{Classification F-Scores with time (secs) for varying Weight Decay rates (LR=1e-3) \label{tab:tablehyperWD}}
  \centering
  \setlength\extrarowheight{-2pt}
   \begin{tabular}  {p{1.8cm} c c c c c c c c c c c c c c c}

  \hline

  \textbf{Model} & WD &  \multicolumn{2}{c}  {\parbox{1.6cm}{\centering NCI1 \\        }} & \multicolumn{2}{c} {\parbox{1.6cm}{\centering Proteins \\        }}  & \multicolumn{2}{c} {\parbox{1.6cm}{\centering Mutag \\        }} & \multicolumn{2}{c} {\parbox{1.6cm}{\centering Cora \\        }}  & \multicolumn{2}{c} {\parbox{1.6cm}{\centering Citeseer \\        }}  & \multicolumn{2}{c} {\parbox{1.6cm}{\centering IMDB-B \\        }} & \multicolumn{2}{c} {\parbox{1.6cm}{\centering REDDIT-B \\        }}  \\

  & & time & f-score & time & f-score & time & f-score & time & f-score & time & f-score & time & f-score & time & f-score \\
  
\hline

{\multirow{1}{*}{\textbf{GCN \cite{kipf2016semi}}}} & 1e-2 & 49.1 &  0.8662 & 19.1 & 0.7805 & 8.6 & 0.4125 & 50.9 & 0.75519  & 49.1 & 0.5133 & 17.2 & 0.7336 & 62.2 & 0.9808 \\

 & 1e-3 & 49.9 & 0.8705 & 18.6 & 0.7806 & 8.3 & 0.4805 & 47.5 & 0.7591 & 48.4 & 0.4925 & 17.4 & 0.7568 & 60.0 & 0.9823 \\
  & 1e-4 & 50.7 & 0.8676 & 18.5 & 0.7808 & 9.0 & 0.4921 & 48.8 & 0.7408  & 49.8 & 0.4688 & 17.2 &  0.7618 & 61.4 & 0.9830 \\

    \hline

 {\multirow{1}{*}{\textbf{GAT \cite{velivckovic2017graph}}}} & 1e-2 & 52.1  & 0.8436 & 19.7 & 0.7906 & 10.1 & 0.5863 & 49.0 & 0.7269  & 49.7 & 0.4962 & 21.2 & 0.7429 & 71.3 & 0.9840 \\

 & 1e-3 & 53.6 & 0.8446 & 19.8 & 0.7945 & 8.4 & 0.6717 & 50.8 & 0.7342  & 49.7 & 0.5030 & 18.4 & 0.7492 & 72.1 & 0.9790 \\
  & 1e-4 & 52.4 & 0.8380 & 19.2 & 0.7898 & 9.0 & 0.6291 & 50.0 & 0.7381  & 49.6 & 0.5071 & 18.0 &  0.7486 & 77.3 & 0.9730 \\
 
    \hline

   {\multirow{1}{*}{\textbf{GIN \cite{xu2018powerful}}}} & 1e-2 & 43.1 & 0.8311 & 16.2 & 0.7742 & 7.5 & 0.6002 & 36.6 & 0.6862  & 34.5 & 0.4468 & 14.9 & 0.6480 & 52.7 & 0.9960 \\

 & 1e-3 & 44.2 &  0.8498 & 16.6 & 0.7794 & 7.6 & 0.6602 & 34.6 & 0.6971 & 33.8 & 0.4660 & 15.0 & 0.6492 & 52.0 & 0.9944 \\
  & 1e-4 & 45.0 & 0.8437 & 16.4 & 0.7797 & 8.1 & 0.6696 & 33.4 & 0.6852  & 34.4 & 0.4365 & 15.6 &  0.6496 & 50.3 & 0.9937 \\

    \hline

     {\multirow{1}{*}{\textbf{GraphSAGE \cite{hamilton2017inductive}}}} & 1e-2 & 47.5 & 0.7447 & 17.8 & 0.7167 &  8.5 & 0.7871 & 40.6 & 0.72172  & 44.2 & 0.4648 & 16.2 & 0.7479 & 41.8 & 0.9738 \\

 & 1e-3 & 48.1 & 0.7670 & 18.9 & 0.7179 & 7.8 & 0.7798 & 41.0 & 0.7137   & 40.7 & 0.4261 & 16.5 & 0.7452 & 44.4 & 0.9558 \\
  & 1e-4 & 49.2 & 0.7598 & 17.6 & 0.7243 & 7.9 & 0.7862 & 41.8 & 0.7222 & 42.5 & 0.4093 & 16.3 & 0.7478 & 44.8 & 0.9717 \\

    \hline

     {\multirow{1}{*}{\textbf{GCN-CAL \cite{sui2022causal}}}} & 1e-2 & 64.3 & 0.8443 & 22.9 & 0.8452 & 10.4 & 0.4034 & 87.2 & 0.7128  & 87.9 & 0.3983 & 22.3 & 0.7970 & 114.5 & 0.9855 \\

 & 1e-3 & 63.4 & 0.8493 & 23.1 & 0.8317 & 10.2 & 0.4296 & 89.5 & 0.6891  & 89.3 & 0.3970 & 22.3 & 0.8113 & 112.9 & 0.9820 \\
  & 1e-4 & 60.9 & 0.8472 & 22.4 & 0.8330 & 10.1 & 0.4474 & 81.0 & 0.6921  & 84.7 & 0.3804 & 22.7 & 0.8084 & 111.4 & 0.9832 \\
 
    \hline

     {\multirow{1}{*}{\textbf{GAT-CAL \cite{sui2022causal}}}} & 1e-2 & 67.2 & 0.8394 & 23.7 & 0.8150 & 10.3 & 0.497 & 79.2 & 0.6702  & 90.6 & 0.3683 & 22.8 & 0.7449 & 130.5 & 0.9841 \\

 & 1e-3 & 69.1 & 0.8387 & 23.7  & 0.8027 & 10.2 & 0.5039 & 79.9 & 0.6593  & 94.0 & 0.3744 & 23.5 & 0.7562 & 128.2 & 0.9788 \\
  & 1e-4 & 67.4  & 0.8445 & 23.7 & 0.8040 & 10.5 & 0.5155 & 93.7 & 0.6616  & 82.2 & 0.3840 & 23.7 & 0.7630 & 127.6 & 0.9821 \\

    \hline

     {\multirow{1}{*}{\textbf{GIN-CAL \cite{sui2022causal}}}} & 1e-2 & 56.1 & 0.8098 & 21.0 & 0.8640 & 10.6 & 0.2498 & 59.2 & 0.6409  & 57.4 & 0.3929 & 20.8 & 0.7592 & 113.6 & 0.9802 \\

 & 1e-3 & 58.3 & 0.8108 & 21.2 & 0.8485 & 10.4 & 0.2458 & 63.3 &  0.6339  & 57.2 & 0.3636 & 21.2 & 0.7513 & 108.7 & 0.9860 \\
  & 1e-4 & 56.3 & 0.8056 & 21.6 & 0.8488 & 9.6 & 0.2188 & 67.1 & 0.6251  & 60.7 & 0.3210 & 20.1 & 0.7627 & 111.0 & 0.9880 \\

    \hline

     {\multirow{1}{*}{\textbf{UHGR-GAT \cite{ding2020unsupervised}}}} & 1e-2 & 1708 & 0.5940 & 514.2 & 0.6118 & 107.6 & 0.6396 & 9.8 & 0.5774  & 10.7 & 0.4230 & 418.2 & 0.4891 & 765.2 & 0.6059 \\

 & 1e-3 & 1719 & 0.5584 & 523.0 & 0.6712 & 111.5 & 0.6811 & 8.9 & 0.6052  & 10.7 & 0.4432 & 416.4 & 0.4871 & 758.6 & 0.5920 \\
  & 1e-4 & 1762 & 0.5445 & 521.4 & 0.6356  & 113.5 & 0.7306 & 9.5 & 0.5808 & 10.2 & 0.4506 & 412.5 & 0.6039 & 781.6 & 0.5504 \\
 
    \hline
     {\multirow{1}{*}{\textbf{UHGR-GCN \cite{ding2020unsupervised}}}} & 1e-2 & 1829 & 0.5900 & 472.6 & 0.6158 & 91.8 & 0.6495 & 9.0 & 0.5116 & 10.1 & 0.4064 & 438.9 & 0.5089 & 786.4 & 0.6039 \\

 & 1e-3 & 1838 & 0.5861 & 443.1 & 0.6752 & 91.4 & 0.7524 & 8.9 & 0.5216  & 9.8 & 0.4268 & 444.2 & 0.5386 & 741.9 & 0.6297 \\
  & 1e-4 & 1793 & 0.5722 & 438.8 & 0.6495 & 93.2 & 0.6633 & 9.1 & 0.5160  & 9.7 & 0.3756 & 451.1 & 0.5069 & 718.2 & 0.6376 \\
  
    \hline

  \end{tabular}
  \label{tab:tablehyperWD}
\end{table*}


\subsubsection{Analysis of training process}

The following findings are observed on analysing the Epochs vs. Accuracy plots for the GCN, GAT, GCN-CAL and GAT-CAL models on the NCI1 and Reddit-B datasets.
\begin{itemize}
    \item In the case of the NCI1 dataset, the GAT model's training and validation scores exhibit a closer alignment, as opposed to the GCN model, indicating good generalisation.
    \item The performance of these four models on the Reddit-B dataset is more consistent, with only the GCN model exhibiting significant variations in training and validation accuracy over the epochs.
    \item The CAL framework displays a more stable validation performance compared to its base model framework. For example, on the NCI1 dataset, the GAT-CAL model exhibits a more consistent performance across epochs compared to the base GAT model. However, this difference is less pronounced in the case of the GAT model on the Reddit-B dataset, possibly due to the diverse nature of the Reddit-B dataset.
    \item In general, the models exhibit variations in performance across datasets; nevertheless, the CAL framework appears to be slightly more consistent and dependable in its overall performance.
    
\end{itemize}

\subsubsection{Analysis of evaluation result} 

Drawing from the evaluation results, the research questions are addressed as follows:

\begin{itemize}
    \item \textbf{RQ1.}  On the NCI1 dataset, the GCN-CAL model demonstrates an additional 1\% accuracy compared to the GAT-CAL model. Similarly, GraphSAGE exhibits slightly less than 1\% extra accuracy over the GAT-CAL model on the Proteins dataset, and GCN outperforms the GAT-CAL model on the IMDB-B dataset by less than 1\%. On the Mutag and Citeseer datasets, the GAT model achieves an extra 3\% accuracy each compared to the GAT-CAL model. Overall, the GAT-CAL architecture consistently delivers excellent performance across all datasets. On the NCI1 dataset, GCN-CAL has shown an extra 1\% accuracy over GAT-CAL model. 
    
    \item \textbf{RQ2.} Within the enhanced GNN architectures, only the GAT-CAL model consistently exhibits high performance across all datasets. Following GAT-CAL, it is noted that the baseline models, viz. GCN and GAT are the top-performing models across all datasets. 
    
    \item \textbf{RQ3.} For node classification tasks, the GAT model, closely followed by the GAT-CAL model, consistently showcases the highest performance.
    
    \item \textbf{RQ4.} The experimental results establish that different learning rates lead to varied performances for all models across all datasets. In contrast, it is shown that varying weight decay rates consistently yields different performance results only for the UHGR models.
    
\end{itemize}

\subsection{Case Studies}

A brief case study was undertaken to assess the models' adaptability to multi-class datasets in the context of graph classification. To achieve this, the multi-class variants of the IMDB and REDDIT datasets are employed. The IMDB-MULTI, REDDIT-MULTI-5K and REDDIT-MULTI-12K datasets consists of 3, 5 and 11 classes respectively. Memory constraints prevent the experimentation of UHGR models on the latter dataset. 

\subsubsection*{Results and Analysis}

\renewcommand{\arraystretch}{2}
\begin{table*}[hbt!]
\fontsize{8pt}{8pt}\selectfont
\caption{Test Accuracy (\%), F-scores with time(secs) for graph classification tasks on multi-class datasets \label{tab:tablStudyCase}}
  \centering
  \setlength\extrarowheight{-2pt}
   \begin{tabular}  {p{3cm} c c c c c c c c c }
  \hline
 
  \textbf{Model} &  \multicolumn{3}{c}  {\parbox{4.3cm}{\centering IMDB-MULTI \\        }} & \multicolumn{3}{c} {\parbox{4.3cm}{\centering REDDIT-MULTI-5K \\        }}  & \multicolumn{3}{c} {\parbox{4.3cm}{\centering REDDIT-MULTI-12K \\        }}   \\

  &  time & accuracy & f-score & time & accuracy & f-score & time & accuracy & f-score  \\
  
\hline
    {\multirow{1}{*}{\textbf{GCN \cite{kipf2016semi}}}}  & 19.6 & 51.00 & 0.6652 & 127.8 & 56.49 & 0.7805 & 258.9 & 49.25 & 0.7302 \\
    
    {\multirow{1}{*}{\textbf{GAT \cite{velivckovic2017graph}}}}  & 20.4 & 50.73 & 0.6722 & 169.7 & 55.85 & 0.7475 & 351.1 & 49.48 & 0.7197 \\

    {\multirow{1}{*}{\textbf{GIN \cite{xu2018powerful}}}}  & 16.2 & 49.33 & 0.7460 & 113.8 & 54.15 & 0.6912 & 245.4 & 48.17 & 0.7268 \\
    
    {\multirow{1}{*}{\textbf{GraphSAGE \cite{hamilton2017inductive}}}}  & 21.6 & 50.80 & 0.6882 & 116.52 & 38.85 & 0.0307 & 197.8 & 34.29 & 0.9247 \\
    
    {\multirow{1}{*}{\textbf{GCN-CAL \cite{sui2022causal}}}}  & 23.1 & 51.00 & 0.6379 & 260.5 & 56.57 & 0.7778 & 494.3 & 49.00 & 0.7483 \\
    
    {\multirow{1}{*}{\textbf{GAT-CAL \cite{sui2022causal}}}}  & 25.7 & 49.33 & 0.6164 & 273.5 & 55.69 & 0.7613 & 675.1 & 49.57 & 0.7320 \\
    
    {\multirow{1}{*}{\textbf{GIN-CAL \cite{sui2022causal}}}}  & 21.1 & 48.73 & 0.6838 & 255.5 & 54.11 & 0.6343 & 502.1 & 47.29 & 0.7044 \\
    
    {\multirow{1}{*}{\textbf{UHGR-GAT \cite{ding2020unsupervised}}}}  & 319.9 & 38.66 & 0.4158 & 935.4 & 31.97 & 0.4356 & - & - & - \\
    
    {\multirow{1}{*}{\textbf{UHGR-GCN \cite{ding2020unsupervised}}}} & 296.3 & 42.66 & 0.4158 & 898.5 & 37.18 & 0.3861 & - & - & - \\

    \hline
    
  \end{tabular}
  \label{tab:tableCaseStudy}
\end{table*}

\begin{figure}[h]
    \centering
    \includegraphics[width=\columnwidth]{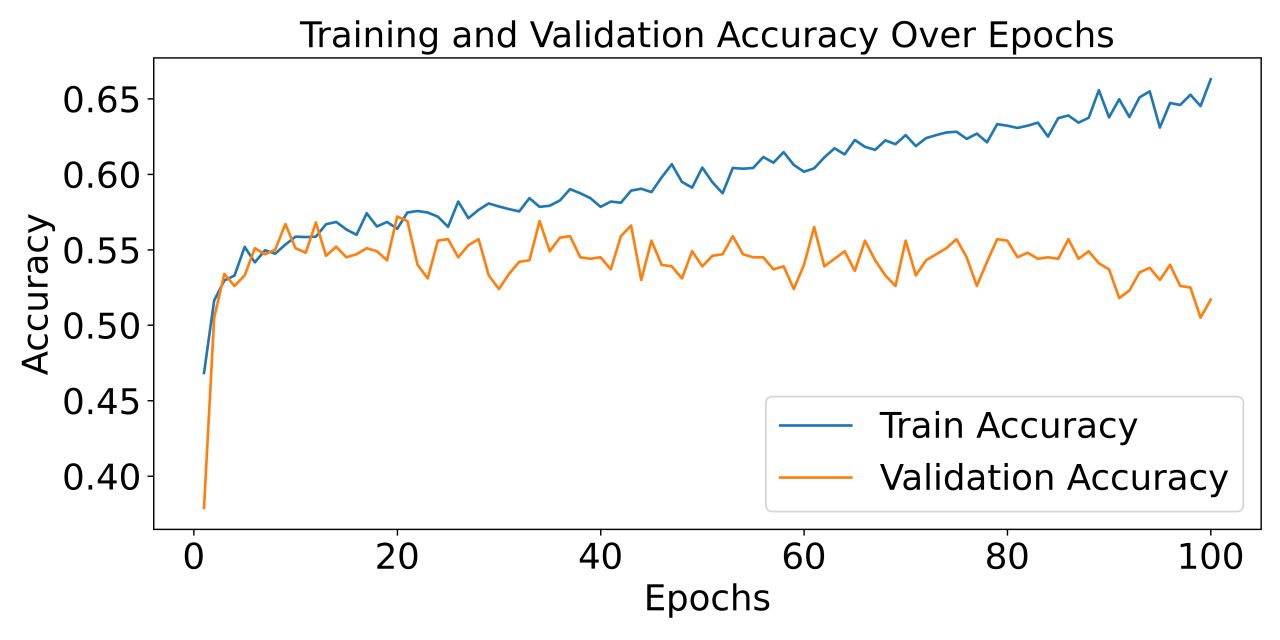}
    \caption{Epochs vs. Accuracy plot for \textit{GAT} model (\textit{Reddit-MULTI-5K} dataset)}
    \label{fig:redditmulti5GAT}
\end{figure}

\begin{figure}[h]
    \centering
    \includegraphics[width=\columnwidth]{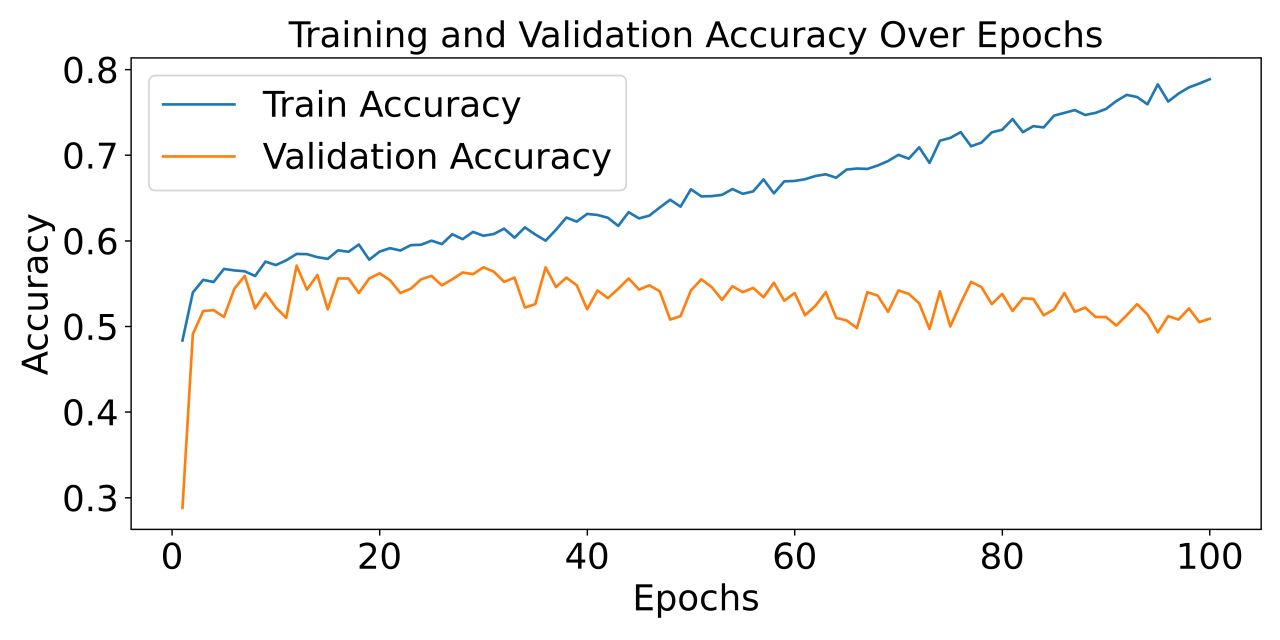}
    \caption{Epochs vs. Accuracy plot for \textit{GAT-CAL} model (\textit{Reddit-MULTI-5K} dataset)}
    \label{fig:redditmulti5GATCAL}
\end{figure}

The experimental configurations remain consistent with the previous settings in our study and the test accuracy and F-Score results are shown in Tab.~\ref{tab:tablStudyCase}. The training plots for the GAT and the GAT-CAL models on the REDDIT-MULTI-5K dataset are shown in Fig.~\ref{fig:redditmulti5GAT} and Fig.~\ref{fig:redditmulti5GATCAL} respectively. Upon examining these findings, the following observations are noted:
\begin{itemize}
    \item In the case of the IMDB-MULTI dataset, the GIN model attains the highest F1-score of 0.7460. However, the corresponding performance for the IMDB-B dataset is lower at 0.6609 when compared to the majority of other GNN models.
    
    \item In the context of the REDDIT datasets, both GCN and GAT, along with their respective CAL architectures, demonstrate comparable high performance levels for both the REDDIT-MULTI-5K and REDDIT-B datasets.
    
    \item A particularly noteworthy finding is the significant disparity in GraphSAGE performance between the REDDIT-MULTI-5K dataset, where it is notably low, and the REDDIT-MULTI-12K dataset, where it surpasses other GNN models to achieve the highest performance. This can be ascribed to the fact that the relationships between nodes in the 5-class dataset may be more intricate or less compatible with GraphSAGE's neighborhood aggregation approach. 

    \item Upon thorough examination, it is evident that the performance outcomes are subpar for the multi-class datasets, with this pattern being particularly  noticeable in the case of the Reddit datasets.
    
    \item On analysing the training plots for the GAT and GAT-CAL models on the REDDIT-MULTI-5K dataset, it is observed that the validation performance deviates significantly away from the training performance, especially when compared to the corresponding 2-class dataset, depicted in figures Fig.~\ref{fig:gat_RedditB} and Fig.~\ref{fig:cal_gat_RedditB} respectively.
\end{itemize}

\subsection{Summary}

On an overall analysis of the results, it is observed that the baseline models of GNNs have, more or less, performed well when compared with the attention-derived causal model, namely the CAL framework. However, the GNN model enabled with MI has shown some shortcomings in these tasks, indicating the need for additional improvements to enhance its performance. Moreover, our research underscores the importance of hyperparameter tuning to achieve improved model performance. Additionally, our findings suggest that the models can perform reasonably well on multi-class datasets in the context of graph classification.


\section{Conclusion} \label{sec-CON}

This paper presented an in-depth study on the application of graph neural networks for potential causal classification. Through a rigorous evaluation, we have analyzed the performance and applicability of standard GNN models, as well as those specifically designed for causality analysis. Our study offers insights into the performance of various GNN architectures and their generalizability on datasets from diverse domains. Additionally, we investigated the effectiveness of causality-based GNN frameworks in node and graph classification tasks, including a sensitivity analysis to understand the impact of hyperparameter variations on model performance. While the study extends to the application in multi-class graph classification, it reveals that reliance on attention mechanisms or mutual information estimation alone does not suffice for accurate causality inference in GNN models. This underscores the necessity for further research and the development of more sophisticated techniques to construct a robust causal GNN classification framework.

\section*{Acknowledgments}
This work is partially supported by The work is partially supported by grants from Australian Research Council (No. DP220101360) and the SAGE Athena Swan Scholarship, UniSQ.

\FloatBarrier
\small
\bibliographystyle{IEEEtranN}
\bibliography{gnnemp}

\end{document}